\documentclass[preprint,12pt,authoryear]{elsarticle}
\usepackage{amssymb}
\usepackage{amsmath}
\usepackage{lineno}
\usepackage{placeins}
\usepackage{booktabs}
\usepackage{nccmath} 
\usepackage{hyperref}
\usepackage{cleveref}

\DeclareMathOperator{\Var}{Var}
\DeclareMathOperator{\Cov}{Cov}

\usepackage{mathtools}

\usepackage[nolist,nohyperlinks]{acronym}
\acrodef{SHAP}{Shapley additive explanations}
\acrodef{SAGE}{Shapley additive global importance}
\acrodef{ERFC}{expected relative feature contribution}
\acrodef{SNP}{single nucleotide polymorphism}
\acrodef{GWAS}{genome-wide association study}

\DeclareMathOperator{\Binom}{Binom}
\DeclareMathOperator{\Poisson}{Poisson}
\DeclareMathOperator{\Unif}{Unif}
\DeclareMathOperator{\N}{N}
\DeclareMathOperator{\shape}{shape}
\DeclareMathOperator{\rate}{rate}
\DeclareMathOperator{\size}{size}
\DeclareMathOperator{\SHAP}{SHAP}
\DeclareMathOperator{\SAGE}{SAGE}
\DeclareMathOperator{\BMI}{BMI}

\makeatletter
\def\ps@pprintTitle{%
  \let\@oddhead\@empty
  \let\@evenhead\@empty
  \let\@oddfoot\@empty
  \let\@evenfoot\@oddfoot
}
\makeatother

\begin{document}
\begin{frontmatter}

\title{
Inferring feature importance with uncertainties in high-dimensional data}

\author[inst1,inst2]{P{\aa}l Vegard Johnsen}
\author[inst3,inst4]{Inga Str{\"u}mke}
\author[inst1]{Signe Riemer-Sørensen}
\author[inst5]{Andrew Thomas DeWan}
\author[inst2]{Mette Langaas}

\affiliation[inst1]{organization={SINTEF DIGITAL},
            postcode={0373}, 
            state={Oslo},
            country={Norway}}
\affiliation[inst2]{organization={Department of Mathematical Sciences, Norwegian University of Science and Technology},
            postcode={7491}, 
            state={Trondheim},
            country={Norway}}
\affiliation[inst3]{organization={Department of Engineering Cybernetics, Norwegian University of Science and Technology},
            postcode={7034}, 
            state={Trondheim},
            country={Norway}}
\affiliation[inst4]{organization={Department of Holistic Systems, SimulaMet},
            postcode={0167}, 
            state={Oslo},
            country={Norway}}
\affiliation[inst5]{organization = {Department of Chronic Disease Epidemiology and Center for Perinatal, Pediatric and Environmental Epidemiology, Yale School of Public Health},
            postcode={CT 06510, New Haven}, 
            state={Connecticut},
            country={USA}}

\begin{abstract}
Estimating feature importance is a significant aspect of explaining data-based models. Besides explaining the model itself, an equally relevant question is which features are important in the underlying data generating process. We present a Shapley value based framework for inferring the importance of individual features, including uncertainty in the estimator. We build upon the recently published feature importance measure of SAGE (Shapley additive global importance) and introduce sub-SAGE which can be estimated without resampling for tree-based models. We argue that the uncertainties can be estimated from bootstrapping and demonstrate the approach for tree ensemble methods. The framework is exemplified on synthetic data as well as high-dimensional genomics data.
\end{abstract}

\end{frontmatter}
\section{Introduction}
With the strong improvement of black-box machine learning models such as gradient boosting models and deep neural networks, the question of how to infer feature importance in these types of models has become increasingly important. 
The Shapley decomposition, a solution concept from cooperative game theory~\citep{shapley_original}, has enjoyed a surge of interest in the literature on explainable artificial intelligence in recent years, (cf. \cite{aas_explaining_2021,lundberg_local_2020,[S4],Lundberg2017,[S6],[S8],[S11],[S12],[S13],[S15],[S16],[S17],[F4],fryer2021model}).
A widely used Shapley based framework for deriving feature importances in a fitted machine learning model is \ac{SHAP}~\citep{Lundberg2017,lundberg_local_2020}, which explains single predictions' deviations from the average model prediction. As such, \ac{SHAP} attributes feature importances as they are perceived by the \textit{model}.
The more recently introduced \ac{SAGE} is also based on the Shapley decomposition, but attributes feature importances by a global decomposition of the model loss across a whole data set~\citep{covert2020understanding}. The \ac{SAGE} framework thus provides an explanation of the influence of the features taking into account not only the model, but also implicitly the data via the loss function, thus encapsulating that the model might not be -- and most likely isn't -- a perfect description of the data~\citep[see][for a discussion and comparison between SHAP and SAGE as feature performance measures]{shapley_good_bad}.

The \ac{SAGE} value needs to be estimated, and the \ac{SAGE} estimator is itself a random variable as the corresponding \ac{SAGE} estimate is based on data of finite size generated from some unknown probability distribution. As is the case for any feature importance measure, we argue that the uncertainty in the estimate is equally important as the estimate itself for drawing conclusions. However, even computation of the \ac{SAGE}-estimate is infeasible for high-dimensional data, and thus further approximations are needed~\citep{covert2020understanding}.  
To this end, we introduce sub-\ac{SAGE}, which is motivated by \ac{SAGE} but can be estimated exactly for tree-ensemble models, by using a reduced subset of coalitions. Additionally, we describe how to estimate a confidence interval of the sub-SAGE value. No calculation of such uncertainty exists in the \ac{SAGE} package or the literature.
We do this using paired bootstrapping, and demonstrate its calculation on simulated as well as observed high-dimensional data. We argue that this procedure provides a way to infer the true importance of a feature in the underlying data. 
We restrict ourselves to tree ensemble models. The remainder of this paper is structured as follows. In~\cref{sec:Background} we introduce background concepts such as the Shapley value, \ac{SHAP} and \ac{SAGE}, before moving on to sub-SAGE in~\cref{sec:subSAGE} and its uncertainty in \cref{sec:inference}. The method is exemplified in~\cref{sec:synthetic} and~\cref{sec:genetic} before we discuss the results in~\cref{sec:discussion}.

\section{Background} \label{sec:Background}
In this section, we provide a brief introduction to the Shapley decomposition-based \ac{SHAP} and \ac{SAGE} frameworks, and how to apply these to tree ensemble models.
The Shapley decomposition is a solution concept from cooperative game theory~\citep{shapley_original}. It provides a decomposition of \textit{any} value function $v(\s)$ that characterises the game, and produces a single real number, or payoff, per set of players in the game. The resulting decomposition satisfies the three properties of efficiency, monotonicity and symmetry, and is provably the only method to satisfy all three~\citep[Thm. 2]{Young:1985aa,Huettner:2012aa}. For details see \ref{app:sage_axiom_properties}.

Consider a supervised learning task characterised by a set of $M$ features $\mathbf{x}_i$ and corresponding univariate\footnote{The procedures described in this paper can be generalised to multivariate responses, but this renders the derivations more convoluted.} responses $y_i$, for $i = 1, \dots, N$, and a fitted model that is a mapping from feature values to response values, i.e.\ $\mathbf{x}_i \rightarrow \hat{y}(\mathbf{x}_i)$. As usual, uppercase letters denote random variables while lowercase letters denote observed data values. In this work, we assume independent features, meaning $E[X_{j}|X_{k} = x_{k}] = E[X_{j}]$ $\forall \ j \ne k$. 
\subsection{The SHAP value}
 Let $\mathcal{S} \subseteq \mathcal{M}\setminus \{k\}$, with $\mathcal{M} = \{1,\ldots,M\}$, denote a subset of all features not including feature $k$. Denote $\bar{\mathcal{S}}$ the corresponding complement subset of excluded features ($\mathcal{S} \cup \bar{\mathcal{S}} = \mathcal{M}$). The \ac{SHAP} value, $\phi^{\SHAP}_k(\textbf{x},\hat{y})$, introduced by~\cite{Lundberg2017}, for a feature with index $k$ with respect to feature values $\textbf{x}$ and a corresponding fitted model $\hat{y}$, is defined as
\begin{equation}
    \phi^{\SHAP}_{k}(\textbf{x},\hat{y}) = \sum_{\mathcal{S} \subseteq \mathcal{M} \setminus \{k\}} \frac{|\mathcal{S}|!(M-|\mathcal{S}|-1)!}{M!}\left[v_{\textbf{x},\hat{y}}(\mathcal{S} \cup \{k\})-v_{\textbf{x},\hat{y}}(\mathcal{S})\right]\,.
\end{equation}
Here, the value function $v_{\textbf{x},\hat{y}}(\s)$ is defined as the expected output of a prediction model conditioned that only a subset $\s$ of all features are included in the model,
\begin{equation}
 v_{\textbf{x},\hat{y}}(\s) =  E_{\textbf{X}_{\overline{\s}}}[\hat{y}(\textbf{X}|\textbf{X}_{\s} = \textbf{x}_{\s})] \,.
 \label{vS}
\end{equation}
For instance, if $\mathbf{x}_{\overline{\s}}$ is continuous and we assume all features to be mutually independent, we have
\begin{align}
\begin{split}
    E_{\mathbf{X}_{\overline{\mathcal{S}}}}[\hat{y}(\mathbf{X} |\mathbf{X}_{{\mathcal{S}}}=\mathbf{x}_{{\mathcal{S}}})] 
    &= \int_{\mathbf{x}_{\overline{\mathcal{S}}}} 
    \hat{y}\left(\mathbf{X}_{\mathcal{S}} = \mathbf{x}_{\mathcal{S}}, \mathbf{X}_{\overline{\mathcal{S}}} = \mathbf{x}_{\overline{\mathcal{S}}}\right)
    p\left(\mathbf{X}_{\overline{\mathcal{S}}} = \mathbf{x}_{\overline{\mathcal{S}}}|\mathbf{X}_{\mathcal{S}} = \mathbf{x}_{\mathcal{S}}\right)
    d\mathbf{x}_{\overline{\mathcal{S}}} \\
    &=\int_{\mathbf{x}_{\overline{\mathcal{S}}}} 
    \hat{y}\left(\mathbf{X}_{\mathcal{S}} = \mathbf{x}_{\mathcal{S}}, \mathbf{X}_{\overline{\mathcal{S}}} = \mathbf{x}_{\overline{\mathcal{S}}}\right) 
    p\left(\mathbf{X}_{\overline{\mathcal{S}}} = \mathbf{x}_{\overline{\mathcal{S}}}\right) d \mathbf{x}_{\overline{\mathcal{S}}} \,.
\end{split}
\label{EfS}
\end{align}
The stochastic behaviour in $\hat{y}(\mathbf{X} |\mathbf{X}_{{\mathcal{S}}}=\mathbf{x}_{{\mathcal{S}}})$ is due to the random vector $\mathbf{X}_{\bar{\mathcal{S}}}$ of unknown feature values. We can think of the difference $v_{\textbf{x},\hat{y}}(\mathcal{S} \cup \{k\})-v_{\textbf{x},\hat{y}}(\mathcal{S})$
as the mean difference in a single model prediction when using feature $k$ in the model compared to when the value of feature $k$ is absent. Therefore, the \ac{SHAP} value can be interpreted as a feature importance measure for each single model prediction. The larger absolute \ac{SHAP} value a feature $k$ has in a single prediction, the more influence the feature is regarded to have.
\subsection{The SAGE value}
Define a loss function $\ell(y_i,\hat{y}(\mathbf{x}_i))$ as a measure of how well the fitted model $\hat{y}(\mathbf{x}_i)$ maps the features to a response, compared to the true response value $y_i$. As defined in~\cite{covert2020understanding}, we take the \ac{SAGE} value function $w(\mathcal{S})$ as the expected difference in the observed value of the loss function when the features in $\mathcal{S}$ are included in the model compared to excluding all features,
\begin{equation}
    w_{\textbf{X},Y,\hat{y}}(\s) = E_{\mathbf{X},Y}[\ell(Y,V_{\textbf{X},\hat{y}}(\emptyset))]-E_{\mathbf{X},Y}[\ell(Y,V_{\textbf{X},\hat{y}}(\s))] \,.
    \label{wS-wphi}
\end{equation}
Here, $\emptyset$ denotes the empty set, while $V_{\textbf{X},\hat{y}}(\s)$ is the stochastic version of \cref{vS}. Specifically, $V_{\textbf{X},\hat{y}}(\s)$ is a random variable since its observed value varies depending on the random vector $X_{\s}$, while $v_{\textbf{x},\hat{y}}(\s)$ is a constant as we condition on the \textit{observed} vector $\textbf{x}_{\s}$.
For instance, for the case where $\textbf{x}$ and $y$ are continuous, the expected value of the loss function when only a subset $\mathcal{S}$ of feature values are known is
\begin{equation}\medmath{  
    E_{\mathbf{X},Y}[\ell(Y,V_{\textbf{X},\hat{y}}(\s))] 
    = \int_{y} \int_{\mathbf{x}_{\mathcal{S}}} \ell \left(y(\textbf{x}),E_{\mathbf{X}_{\bar{\mathcal{S}}}}\left[\hat{y}\left(\mathbf{X} |\mathbf{X}_{{\mathcal{S}}} 
    = \mathbf{x}_{\mathcal{S}} \right)\right]\right) p(y|\textbf{x}_{\s})p(\mathbf{x}_{\mathcal{S}})d\mathbf{x}_{\mathcal{S}}dy} \,.
\label{wS}
\end{equation}
Notice that the computation of 
$v_{\textbf{x},\hat{y}}(\s) = E_{\mathbf{X}_{\bar{\mathcal{S}}}}\left[\hat{y}\left(\mathbf{X} |\mathbf{X}_{{\mathcal{S}}} 
= \mathbf{x}_{\mathcal{S}} \right)\right]$ 
happens inside the loss function, which is usually non-linear. Also notice that in \cref{wS}, we integrate over \textit{all} possible values of $X_{\s}$.

The \ac{SAGE} value for a feature $k$ is defined as
    \begin{equation}
    \phi^{\SAGE}_{k}(\textbf{X},Y,\hat{y}) = \sum_{\mathcal{S} \subseteq \mathcal{M} \setminus \{k\}} \frac{|\mathcal{S}|!(M-|\mathcal{S}|-1)!}{M!}\left[w_{\textbf{X},Y,\hat{y}}(\mathcal{S} \cup \{k\})-w_{\textbf{X},Y,\hat{y}}(\mathcal{S})\right] \,.
    \label{sageDef}
\end{equation}
We can think of the difference $w_{\textbf{X},Y,\hat{y}}(\mathcal{S} \cup \{k\})-w_{\textbf{X},Y,\hat{y}}(\mathcal{S})$ as the expected difference in the loss function when including feature $k$ in the model compared to excluding feature $k$ with respect to the subset $\s$ of known feature values. \ac{SAGE} is therefore a global feature importance measure, as opposed to the \ac{SHAP} value, as it does not evaluate a single prediction, but rather the impact feature $k$ has across all predictions. The use of the loss function in the \ac{SAGE} definition also makes sure that the feature importance is not only based on the model, as for the \ac{SHAP} value, but also on the data itself.
    
The features and response can be both continuous and discrete. In the discrete case, integrals must replaced by sums and vice versa in~\cref{EfS,wS}. The expressions in~\cref{vS,wS-wphi} are in general unknown and need to be estimated for each choice of model and loss function. Consequently, the \ac{SHAP} and \ac{SAGE} values become estimates as well.

An interpretation of \ac{SAGE} is that a positive \ac{SAGE} value for a features implies that including this feature in the model reduces the expected model loss compared to when not including the feature. 
\subsection{Tree ensemble models}
\begin{figure}
    \centering
    \includegraphics[width=0.7\textwidth]{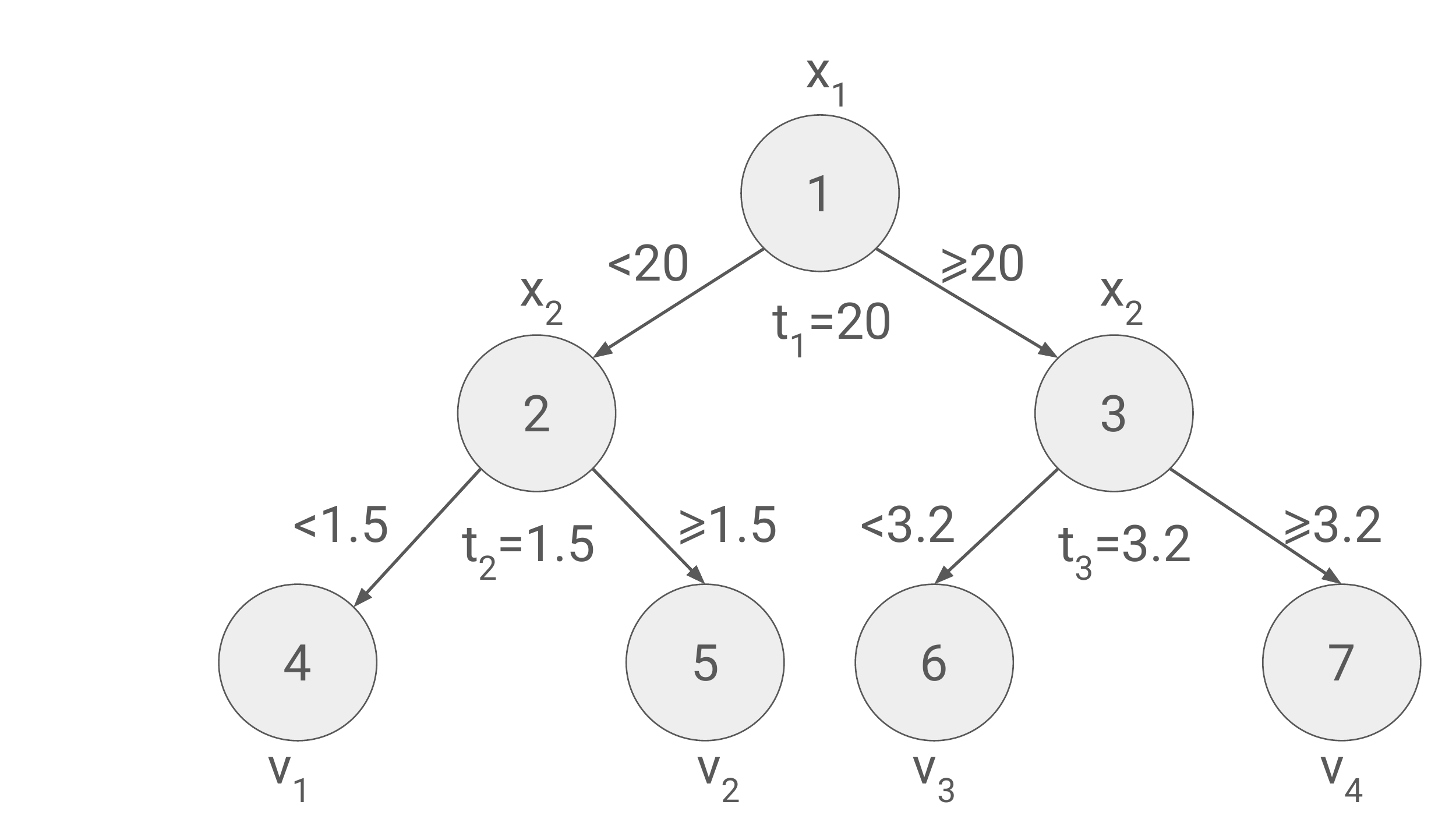}
    \caption{\label{exampleRegtree}A regression tree including two features $X_1$ and $X_2$.}
\end{figure}

Consider a tree ensemble model consisting of several regression trees $f_{\tau}(\mathbf{x}_i)$ with predicted response $\hat{y}(\mathbf{x}_i)$, such that $\hat{y}(\mathbf{x}_i) = \sum_{\tau=1}^T f_{\tau}(\mathbf{x}_i)$ for $T$ trees. By the linearity property of the expected value, we have
\begin{equation}
    v_{\textbf{x},\hat{y}}(\s) 
    = E_{\mathbf{X}_{\bar{\mathcal{S}}}}\left[\sum_{\tau =1}^T f_{\tau}(\mathbf{X} |\mathbf{X}_{{\mathcal{S}}} 
    = \mathbf{x}_{{\mathcal{S}}} )\right] = \sum_{\tau = 1}^T E_{\mathbf{X}_{\bar{\mathcal{S}}}}[f_{\tau}(\mathbf{X} |\mathbf{X}_{{\mathcal{S}}} = \mathbf{x}_{{\mathcal{S}}} )] \,.
\end{equation}
The computation of $E_{\mathbf{X}_{\bar{\mathcal{S}}}}[f_{\tau}(\mathbf{X} |\mathbf{X}_{{\mathcal{S}}} = \mathbf{x}_{{\mathcal{S}}} )]$ can be understood through a simple example: Consider the regression tree illustrated in~\cref{exampleRegtree}. It has depth two and splits on the two features indexed 1 and 2, which are continuous and mutually independent. The regression tree has parameters such as \textit{splitting points}, $t_j$, for branch nodes, and \textit{leaf values} $v_j$, for leaf nodes. Assume that $x_2=3$ is observed. We then have
\begin{align}
    \begin{split}
    E_{\mathbf{X}_{\bar{\mathcal{S}}}}[f_{\tau}(\mathbf{X} |\mathbf{X}_{{\mathcal{S}}} = \mathbf{x}_{{\mathcal{S}}} )] 
    & = E_{X_1}[f_{\tau}(X_1 |X_2 = 3 )] \\
    & = P(X_1 \ge 20) v_3 + P(X_1 < 20) v_2 \,.
    \label{exampleexpectedtree}
    \end{split}
\end{align}
In general, we do not know the value of $P(X_1 \le 20)$, and need to estimate it. Consider $N$ data instances with recorded feature values from feature $k$. An \textit{unbiased} estimate of $P(X_k \le t)$ is then
\begin{equation}
    \hat{P}(X_k \le t) = \frac{1}{N} \sum_{i=1}^N I(x_{i,k} \le t) \,,
\label{unbiasedEstimate}
\end{equation}
where $x_{i,k}$ is the observed value of feature $k$ for data instance $i$. Using this estimate, we can also get an unbiased estimate for~\cref{exampleexpectedtree}. An unbiased estimate of $E_{\mathbf{X}_{\bar{\mathcal{S}}}}[f_{\tau}(\mathbf{X} |\mathbf{X}_{{\mathcal{S}}} = \mathbf{x}_{{\mathcal{S}}} )]$ for any regression tree can be achieved by a recursive algorithm~\citep{lundberg_local_2020} with running time $O(L 2^{M})$, where $L$ is the number of leaves, see~\cref{RecAlg}.

\begin{algorithm}
	\caption{Recursive algorithm for computation of $E_{\mathbf{X}_{\bar{\mathcal{S}}}}[f_{\tau}(\mathbf{X} |\mathbf{X}_{{\mathcal{S}}} = \mathbf{x}_{{\mathcal{S}}} )]$.} 
	\begin{algorithmic}[1]
	    \State Input: Tree $f_{\tau}$ with depth $d$, leaf values $\textbf{v} = (v_1,\ldots,v_{2^{d}})$, feature used for splitting $\textbf{f} = (f_1,\ldots,f_{2^{d}-1})$ and corresponding splitting points $\textbf{t}=(t_1,\ldots,t_{2^{d}-1})$. Estimated probabilities  of ending at a node $j$ given previous information, for all nodes in the tree, $\textbf{p} = (p_1,\ldots,p_{2^{d}-1})$, by using some data $(\textbf{x}_1,y_1),\ldots,(\textbf{x}_N,y_N)$ of size $N$. The subset of features $\mathcal{S}$ with corresponding known values $x_{\s}$. The left and right descendant node for each internal node $\textbf{l} = (l_1,\ldots,l_{2^{d}-1})$ and $\textbf{r} = (r_1,\ldots,r_{2^{d}-1})$. The index of a node $j$ in the tree $f_{\tau}$.
	    \State \textbf{Function} CondExpTree($j,f_{\tau},\bf{v},\bf{t},\bf{f},\bf{l},\bf{r},\bf{p}$)
	    \If{IsLeaf(j)}
	        \State return $v_j$
	    \Else
	        \If{$f_j \in \s$}
	            \If{$x_j \le t_j$}
	                \State return CondExpTree($l_j,f_{\tau},\bf{v},\bf{t},\bf{f},\bf{l},\bf{r},\bf{p}$)
	            \Else 
	                \State return CondExpTree($r_j,f_{\tau},\bf{v},\bf{t},\bf{f},\bf{l},\bf{r},\bf{p}$)
	            \EndIf
	       \Else
	            \State return CondExpTree($l_j,f_{\tau},\bf{v},\bf{t},\bf{f},\bf{l},\bf{r},\bf{p}$) $p_{l_j}$ + \State \ \ \ \ \ \ \ \ \ CondExpTree($r_j,f_{\tau},\bf{v},\bf{t},\bf{f},\bf{l},\bf{r},\bf{p}$) $p_{r_j}$
	        \EndIf
	    \EndIf
		\State \textbf{End Function}
		
		\State CondExpTree($1,f_{\tau},\bf{v},\bf{t},\bf{f},\bf{l},\bf{r},\bf{p}$) \Comment{Start at root node.}
	\end{algorithmic} 
\label{RecAlg}
\end{algorithm}

\subsection{SAGE in practice}
In practice, as the expressions in \cref{vS} and \cref{wS-wphi} must be estimated, we get a \ac{SAGE} estimator rather than a \ac{SAGE} value. However, since the \ac{SAGE} estimator requires summing over all $2^{M-1}$ subsets $\s \subseteq \mathcal{M} \setminus \{k\}$, for \textit{each} feature, computing the \ac{SAGE} estimator for observed data with many features becomes infeasible. In~\cite{covert2020understanding}, the \ac{SAGE} estimate is approximated through a Monte Carlo simulation process. Specifically, instead of iterating over all $2^{M-1}$ subsets, a subset $\s$ is randomly sampled with replacement in each iteration out of $I$ iterations in total. The differences $w_{\textbf{X},Y,\hat{y}}(\mathcal{S} \cup \{k\})-w_{\textbf{X},Y,\hat{y}}(\mathcal{S})$ for each $\s$ are estimated by sampling data instances with replacement and computing sample means~\cite[see][Appendix D for details]{covert2020understanding}. For an arbitrarily large data set, the authors show convergence to the true \ac{SAGE} estimate as $I \rightarrow \infty$. Among other things, both the accuracy and convergence speed of the algorithm naturally depends on the number of features in the prediction model.

Keeping in mind that the \ac{SAGE} \textit{estimator} is a random variable, we argue that its uncertainty is equally important as the estimate itself. No calculation of this inherent uncertainty exists in the \ac{SAGE} package or the literature \footnote{\cite{covert2020understanding} provides the degree of convergence of the approximation of the estimate, not the uncertainty in the estimate.}.
To this end, we introduce \textit{sub-\ac{SAGE}}, which is inspired by the \ac{SAGE} framework, but consisting of a reduced number of subsets $\s \in \mathcal{Q}$. While applicable to any number of features, it is best suited for interpreting a small number of features, or a small subset of features in a large feature set. 

\section{Sub-SAGE} \label{sec:subSAGE}
Given hundreds or thousands of features in a model, the computation time required to get a satisfactory accurate estimate of \ac{SAGE}~\citep{covert2020understanding}, for each feature, quickly becomes unacceptable. A hybrid approach is to select a reduced subset of features of particular interest to investigate. Such a subset can for instance be selected by computing a model-based feature importance score for all features in the model and selecting the most interesting looking ones. The reduced subset of promising features can then by more thoroughly investigated in order to infer whether their model-based importance is also reflected in the underlying data generating process. For this purpose, we introduce \textit{sub-\ac{SAGE}}, where only a selection of the in total $2^{M-1}$ subsets are involved in the computation of each feature.

If we want to measure the importance of a feature $k$ based on its marginal effect, as well as potential pairwise interactions it may be involved in, computing $\mathcal{S} = \{\emptyset\}$ and $\mathcal{S} = \{m\}$ for $m = 1,\dots,k-1,k+1,\dots,M$ is sufficient. In addition, by including $\mathcal{S} = \{1,\dots,k-1,k+1,\dots,M\}$, the set of all features except feature $k$, this can be used to measure the importance of feature $k$ in the presence of all features at the same time. Let $\mathcal{Q}_k$ denote the set of subsets $\mathcal{S}$ chosen above. We define the sub-SAGE value, $\psi_k$, for feature $k$ as
\begin{equation}
    \psi_{k}(\textbf{X},Y,\hat{y}) = \sum_{\mathcal{S} \in \mathcal{Q}_k} \frac{|\mathcal{S}|!(M-|\mathcal{S}|-1)!}{3(M-1)!}\left[w_{\textbf{X},Y,\hat{y}}(\mathcal{S} \cup \{k\})-w_{\textbf{X},Y,\hat{y}}(\mathcal{S})\right] \,,
    \label{RedSAGEDef}
\end{equation}

Each subset is weighted such that the sum of the weights of all subsets with equal size is the same for each subset size. In addition, the sum of all weights is equal to one. Hence, the construction is similar to the weights defined for Shapley values. See \ref{app:subsage_weights} for details. In this particular case, there are three possible subset sizes, and so the sum of the weights for each subset size is $\tfrac{1}{3}$. Shapley properties such as symmetry, dummy property and monotonicity still holds for sub-SAGE. However, as the sum is not over all possible subsets, the sub-SAGE values do no longer satisfy the efficiency axiom of the Shapley decomposition, which \ac{SHAP} and \ac{SAGE} do (see \ref{app:sage_axiom_properties}) However, we regard the efficiency property as not necessary in this particular setting, as we still consider the sub-SAGE to be informative with respect to feature importance via the computed differences $w_{\textbf{X},Y,\hat{y}}(\mathcal{S} \cup \{k\})-w_{\textbf{X},Y,\hat{y}}(\mathcal{S})$. In addition, the purpose is only to evaluate a small fraction of all features, not all of them. By only considering a reduced number of subsets $\s$, compared to \ac{SAGE}, and only considering a reduced number of features to evaluate, both computing the sub-SAGE estimate as well as the uncertainty in the corresponding sub-SAGE estimator become feasible for black-box models, such as for tree ensemble models as discussed in~\cref{sec:inference}.

\subsection{Using sub-SAGE to infer true relationships in the data}
As the goal is to infer feature importance from a black-box model using sub-\ac{SAGE} values, similar to calculating p-values without taking into account the effect of model selection, we must be extra careful. Any model selection procedure using training data is likely to overfit, resulting in a model containing false relationships that are not a general property of the population from which the data was sampled. It is therefore essential that the sub-SAGE value is calculated using independent data the model was not fitted on. We denote such independent data as test data, $(\mathbf{X}^{0}_1,Y^{0}_1),\dots,(\mathbf{X}^{0}_{N_I},Y^{0}_{N_I})$, with $N_{I}$ samples in total.

Consider a fitted linear regression model $\hat{y}_i = \hat{\bm \beta}^T \bm x_i$. By using test data independent of the data used for constructing the linear regression model, and using the squared error loss, one can show that for a feature $k$, and any $\s \in \mathcal{Q}_k$ (see \ref{linregApp}):

\begin{align}
\begin{split}
&w_{\textbf{X},Y,\hat{y}}(\suk)-w_{\textbf{X},Y,\hat{y}}(\s)= 2\hat{\beta}_k \Cov(Y,X_k) - \hat{\beta}^2_{k}\Var(X_k).
\end{split}
\label{linreg}
\end{align}
As the expression is independent of the subset $\s$, this is also equal to the sub-SAGE value of feature $k$.

The first term in \cref{linreg} can be interpreted as the extent to which the influence of feature $k$ based on the model, constructed using training data, is reflected in the independent test data. If the signs of $\hat{\beta}_k$ and $\Cov(Y,X_k)$ are identical, the first term is positive. If they differ, the sub-SAGE value will always be negative since the second term in \cref{linreg} is always negative. The second term $\hat{\beta_k}^2 \Var(X_k)$ is equal to the increased variance in the model by including feature $k$. So, if the model regards the feature as important (resulting in non-zero $\hat{\beta}_k$), while the covariance between $X_k$ and $Y$ from the independent test data goes in the same direction (same sign as $\hat{\beta}_k$), however small, then the benefit of including feature $k$ in the model is smaller, the larger the variance of the feature, and at some point disadvantageous for sufficiently large variance.

\subsection{sub-SAGE applied on tree ensemble models}
\ac{SHAP} values can be shown to be estimated efficiently for tree ensemble models, even with hundreds of thousands of features \citep[e.g.][]{johnsen_new_2021}, by improving \cref{RecAlg} to get a significantly reduced running time of $O(TLD^2)$, for $T$ trees each of tree depth $D$ \citep[see][for details]{lundberg_local_2020}. Unfortunately, there is no similar way to reduce the running time for estimation of SAGE values, as well as sub-SAGE values, for tree ensemble models with non-linear choices of loss functions~\citep{lundberg_local_2020}.

We consider a tree ensemble model consisting of $T$ trees. Consider a particular feature $k$ to compute the sub-SAGE value as well as a subset $\s \in \mathcal{Q}_k$. We separate the trees in the model into two groups $\tau_{k}$ and the complement group ($\overline{\tau}_k$) where $\tau_k$ is the set of trees including feature $k$ as a splitting feature. The loss function is taken to be the squared error between the response and prediction per sample, i.e.\ $\ell={(y(\textbf{x})-\hat{y}(\textbf{x}))}^2$. Then one can show that (see \ref{app:SAGE_trees} for the derivation),
\begingroup\makeatletter\def\f@size{9}\check@mathfonts
\begin{align}
\begin{split}
    &w_{\textbf{X},Y,\hat{y}}(\mathcal{S} \cup \{k\})-w_{\textbf{X},Y,\hat{y}}(\mathcal{S})  \\
    &=  E_{\mathbf{X},Y} \left[\left(Y\left(\textbf{X}\right)-V_{\textbf{X},\hat{y}}\left(\mathcal{S}\right)\right)^2\right] - E_{\mathbf{X},Y}[{\left(Y\left(\textbf{X}\right)-V_{\textbf{X},\hat{y}}(\mathcal{S} \cup \{k\})\right)}^2] \\
    &=  E_{\mathbf{X},Y}\left[
    2Y\left(\textbf{X}\right)\left(\sum_{j \in \tau_k} V_{\textbf{X},f_j}(\mathcal{S} \cup \{k\})
    - V_{\textbf{X},f_j}(\mathcal{S})\right) 
    + \left(\sum_{j \in \tau_k} V_{\textbf{X},f_j}(\mathcal{S})\right)^2 \right.  \\ 
    &\left. - \left(\sum_{j \in \tau_k} V_{\textbf{X},f_j}(\mathcal{S} \cup \{k\})\right)^2 + 2\left(\sum_{j \notin \tau_k} V_{\textbf{X},f_j}(\mathcal{S})\right)
    \left(\sum_{j \in \tau_k} V_{\textbf{X},f_j}(\mathcal{S} \cup \{k\})-V_{\textbf{X},f_j}(\mathcal{S})\right)\right]\,.
\end{split}    
\end{align}
\endgroup
A commonly used loss function for binary classification problems is binary cross-entropy, $\ell = -y(\textbf{x}) \log \hat{y}(\textbf{x}) -(1-y(\textbf{x}))\log(1-\hat{y}(\textbf{x})) = (1-y(\textbf{x}))\sum_{j=1}^{T} f_{j}(\textbf{x})$ 
$+ \log \left(1+e^{-\sum_{j=1}^{T} f_{j}(\textbf{x})}\right)$. For this loss function, one can show that (see~\ref{app:SAGE_trees})
\begingroup\makeatletter\def\f@size{10}\check@mathfonts
\begin{align}
    \begin{split}
        &w_{\textbf{X},Y,\hat{y}}(\mathcal{S} \cup \{k\})-w_{\textbf{X},Y,\hat{y}}(\mathcal{S}) \\
        &=E_{\textbf{X},Y} \left[\left(1-Y\left(\textbf{X}\right)\right)\sum_{j =1}^{T} V_{\textbf{X},f_j}(\mathcal{S}) + \log \left(1+\exp \left(-\sum_{j=1}^{T} V_{\textbf{X},f_j}(\mathcal{S})\right)\right) \right] \\
        &-E_{\textbf{X},Y}\left[\left(1-Y \left(\textbf{X}\right)\right)\sum_{j = 1}^{T} V_{\textbf{X},f_j}(\mathcal{S} \cup \{k\}) + \log \left(1+\exp \left(-\sum_{j=1}^{T} V_{\textbf{X},f_j}(\mathcal{S} \cup \{k\})\right)\right)\right] \\
        &= E_{\textbf{X},Y} \left[ \left(1-Y \left(\textbf{X}\right)\right) \left(\sum_{j \in \tau_{k}} V_{\textbf{X},f_j}(\mathcal{S}) - V_{\textbf{X},f_j}(\mathcal{S} \cup \{k\}) \right) \right. \\
        &\left. \quad + \log \left( \frac{1+\exp \left(-\sum_{j \in \tau_k} V_{\textbf{X},f_j}(\mathcal{S}) - \sum_{j \notin \tau_k} V_{\textbf{X},f_j}(\mathcal{S}) \right)}{1+\exp \left(-\sum_{j \in \tau_k} V_{\textbf{X},f_j}(\mathcal{S} \cup \{k\}) - \sum_{j \notin \tau_k} V_{\textbf{X},f_j}(\mathcal{S} \cup \{k\}) \right)} \right) \right]\,.
    \end{split}
    \label{Classification}
\end{align}
\endgroup
\subsubsection{Plug-in estimates}
As discussed earlier, the expression $w_{\textbf{X},Y,\hat{y}}(\mathcal{S} \cup \{k\})-w_{\textbf{X},Y,\hat{y}}(\mathcal{S})$  needs to be estimated for each $\s \in \mathcal{Q}_k$, \textit{and} based on data, $(\mathbf{x}^{0}_1,y^{0}_1),\dots,(\mathbf{x}^{0}_{N_I},y^{0}_{N_I})$, never used during training of the model. Let $\hat{v}_{\textbf{x}^{0},y^{0},f_\tau}(\mathcal{S})$ for a particular observation $(\textbf{x}^{0},y^{0})$ and regression tree $f_{\tau}$ denote the estimate of $v_{\textbf{x}^{0},f_{\tau}}(\mathcal{S}) = E_{\mathbf{X}_{\bar{\mathcal{S}}}}[f_{\tau}(\mathbf{X}^{0} |\mathbf{X}^{0}_{{\mathcal{S}}} = \mathbf{x}^{0}_{{\mathcal{S}}} )] $ as described in \cref{RecAlg}. A plug-in \textit{estimate} of $\psi_k$, denoted $\hat{\psi}_k$, for a regression problem with continuous response, for a tree ensemble model using the squared error loss is given by
\small{
\begin{align}
\begin{split}
&\hat{\psi}_{k} = \sum_{\mathcal{S} \in \mathcal{Q}} \frac{|\mathcal{S}|!(M-|\mathcal{S}|-1)!}{3(M-1)!}\left[
\frac{2}{N_I} \sum_{i=1}^{N_I} y^{0}_i \left(\sum_{j \in \tau_k} \hat{v}_{\textbf{x}^{0}_i,f_j}(\suk)-\hat{v}_{\textbf{x}^{0}_i,f_j}(\mathcal{S})\right) \right. \\
&\left. + \frac{1}{N_I}\sum_{i=1}^{N_I}\left(\sum_{j \in \tau_k} \hat{v}_{\textbf{x}^{0}_i,f_j}(\mathcal{S})\right)^2 - \frac{1}{N_I}\sum_{i=1}^{N_I}\left(\sum_{j \in \tau_k} \hat{v}_{\textbf{x}^{0}_i,f_j}(\suk)\right)^2 \right. \\
&\left. + \frac{2}{N_I} \sum_{i=1}^{N_I}\left(\sum_{j \notin \tau_k} \hat{v}_{\textbf{x}^{0}_i,f_j}(\mathcal{S}) \right)\left(\sum_{j \in \tau_k} \hat{v}_{\textbf{x}^{0}_i,f_j}(\suk)- \hat{v}_{\textbf{x}^{0}_i,f_j}(\mathcal{S})\right) \right]\,.
\label{sageestlinreg} 
\end{split}
\end{align}}

The corresponding plug-in estimate for the binary cross-entropy loss given in \cref{Classification} can be found in a similar fashion, basically by estimating expected values as their corresponding sample means. For tree ensemble models with tree stumps (maximum depth of one for each tree), the estimate in \eqref{sageestlinreg} is further reduced and can be expressed as sample variance and covariance terms, see~\ref{app:tree_stumps}.  

\section{Inference of sub-SAGE via bootstrapping} \label{sec:inference}
The importance of any feature may be evaluated by estimating sub-SAGE values. Similar to \ac{SAGE}, a positive sub-SAGE value for a feature $k$ indicates that including the feature in the model is expected, based on the subsets $\s \in \mathcal{Q}_k$, to reduce the loss function. However, the corresponding sub-SAGE plug-in \textit{estimator} given the data generating process $(\mathbf{X}^{0}_1,Y^{0}_1),\dots,(\mathbf{X}^{0}_{N_I},Y^{0}_{N_I})$ from some unknown probability distribution includes uncertainty, and this should be evaluated before making any assumptions about feature importance. The complexity of the sub-SAGE plug-in estimators makes paired bootstrapping a tempting approach. Specifically, the procedure is to iteratively, given independent data points at hand $(\mathbf{x}^{0}_1,y^{0}_1),\dots,(\mathbf{x}^{0}_{N_I},y^{0}_{N_I})$, resample the data points \textit{with replacement} to get a new bootstrapped sample $(\mathbf{x}^{*}_1,y^{*}_1),\dots,(\mathbf{x}^{*}_{N_I},y^{*}_{N_I})$. For each bootstrapped sample, a corresponding plug-in estimate, $\hat{\psi}^{*}_b$, can be computed, and after $B$ iterations, the sample $(\hat{\psi}^{*}_1,\dots,\hat{\psi}^{*}_{B})$ can approximate $B$ realizations arising from the true distribution of the plug-in estimator. A $1-2\alpha$ confidence interval can be approximated by the \textit{percentile interval} given by $[\hat{\psi}^{* (\alpha)},\hat{\psi}^{* (1-\alpha)}]$, where $\hat{\psi}^{* (\alpha)}$ is the $100 \alpha$ empirical percentile, meaning the $B \cdot \alpha$th least value in the ordered list of the samples $(\hat{\psi}^{*}_1,\dots,\hat{\psi}^{*}_{B})$\footnote{Assuming $B \cdot \alpha$ is an integer. See for instance~\cite{efron} for conventions.}. The accuracy in the percentile interval increases for larger number of bootstrap iterations. A typical number is $B = 1000$ regarded to be sufficient in most cases. The algorithm of the paired bootstrap applied specifically to tree ensemble models is given in~\cref{PairsBootstrap}. Notice that for each bootstrap sample, the probability estimates in the trees need to be updated according to~\cref{unbiasedEstimate}. In situations where the plug-in estimator is biased, or there is skewness in the corresponding distribution, the bias-corrected and accelerated bootstrap, first introduced in \cite{efron_better_1987}, may give even more accurate confidence intervals at the cost of considerable increase in computational efforts.
\begin{algorithm}
	\caption{Paired bootstrap of sub-\ac{SAGE} value with percentile interval} 
	\begin{algorithmic}[1]
	    \State Given independent test data $(\mathbf{x}^{0}_1,y^{0}_1),\dots,(\mathbf{x}^{0}_{N_I},y^{0}_{N_I})$, model $\hat{y}(\mathbf{x}) = \sum_{\tau=1}^{T} f_{\tau} (\mathbf{x})$, feature $k$, a loss function and $\alpha$ to estimate $1-2\alpha$ confidence interval:  
	    \State Preallocate vector BootVec of length $B$, the total number of bootstrap iterations.
		\For {$b=1,2,\ldots,B$}
			\State Resample data $N_I$ times with replacement to get 
			\State $(\mathbf{x}^{*}_1,y^{*}_1),\ldots,(\mathbf{x}^{*}_{N_I},y^{*}_{N_I})$
			\State Update probabilities estimates in all the trees in $\hat{y}(\textbf{x})$ to get $\textbf{p}^{*}$
			\State BootVec[b] = $\hat{\psi}^{*}_{k}$
		\EndFor
		\State Percentile interval given by $[\hat{\psi}^{* (\alpha)},\hat{\psi}^{* (1-\alpha)}]$
	\end{algorithmic} 
\label{PairsBootstrap}
\end{algorithm}
\section{Proof of concept - With known underlying data generating process} \label{sec:synthetic}
In this section, we exemplify the sub-\ac{SAGE} method on synthetic data with a known relationship defined as
\begin{align}
    \begin{split}
      f(\mathbf{X}_i) &= a_0 + a_1 X_{i,1} +a_2 X_{i,2} + a_{21} X_{i,1} e^{X_{i,2}} + a_3 X_{i,3}^2 + a_4 \sin(X_{i,4}) \\
        & a_5 \log(1+X_{i,5}) - X_{i,5} I(X_{i,6}>7) + \epsilon_i \, ,
    \end{split}
    \label{SimRegEx}
\end{align}
with $a_0=-0.5, a_1=0.03, a_2=-0.05, a_{21}=0.3, a_3=0.02, a_4=0.35, a_5=-0.2$, and where the features are sampled from the following distributions
\begin{align}
    \begin{split}
        &X_{1} \sim \Binom(\size = 2, p = 0.4) \\
        &X_{2} \sim \Binom(\size=2, p = 0.04) \\
        &X_{3} \sim \Gamma(\shape = 10,\rate = 2) \\
        &X_{4} \sim \Unif(0,\pi) \\
        &X_{5} \sim \Poisson(\lambda = 15) \\
        &X_{6} \sim \N(\mu = 0, \sigma = 10) \\
        &\epsilon_i \sim \N(\mu = 0, \sigma = 2) \,.
    \end{split}
\end{align}
In addition, we generate 94 noise variables. $j = 7,\dots,47$ with a normal distribution $X_{j} \sim \N(\mu_j,\sigma_{j})$ and $j = 48,\dots,100$ with a binomial distribution $X_{j} \sim \Binom(2,p_j)$ where $\mu_j, \sigma_{j}$ and $p_j$ are sampled from a uniform distribution. Data is generated to give a total of $16 000$ samples, and then separated randomly in three disjoint subsets: Data for training $(50\%)$, data for evaluation during training $(30\%)$ and independent test data $(20\%)$ used for estimating sub-\ac{SAGE} values.
We fit an ensemble tree model using XGBoost \citep{xgboost} to the true influential features $1, \dots,6$ together with the noise variables $7, \dots,100$.

The hyperparameters are fixed to $\text{max\_depth}=2$, learning rate $\eta = 0.05$, $\text{subsample}=0.7$, regularization parameters $\lambda = 1$, $\gamma = 0$ and $\text{colsample\_bytree}=0.8$ with $\text{early\_stopping\_rounds}=20$ using training data $(n=8000)$ and validation data $(n=4800)$. See~\citep{xgboost} for details about the hyperparameters. We apply the squared error loss during training. This results in a final model including a total of $230$ trees and $62$ unique features out of the $100$ input-features.

\begin{figure}[b!]
    \centering
    \includegraphics[width=1.0\textwidth]{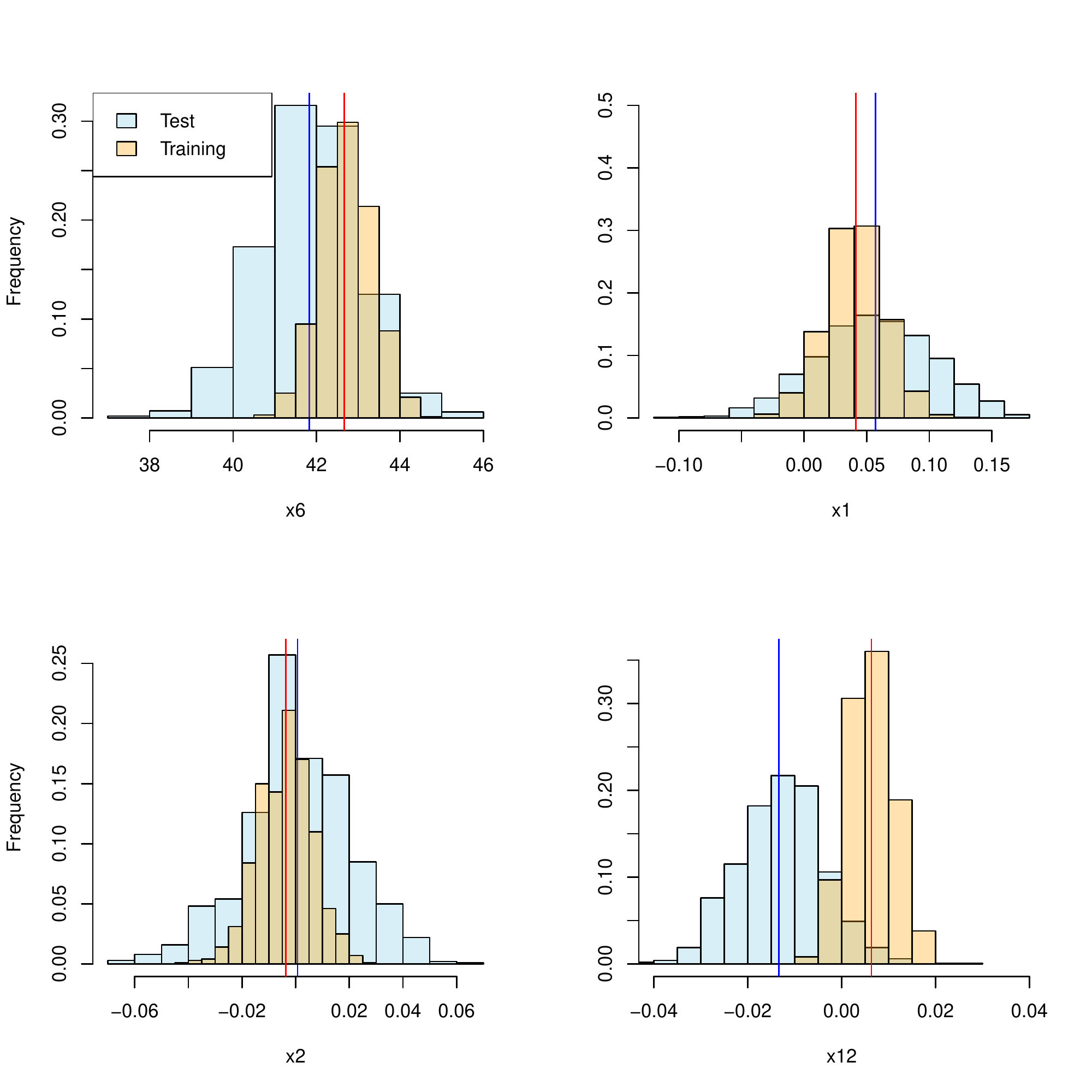}
    \caption{\label{Boothists} The estimate of the sub-SAGE, and the corresponding bootstrap distribution for the synthetic data for features $x_6$, $x_2$, $x_1$ and $x_{12}$, when applying data used during training (orange), and independent test data (blue).}
\end{figure}

From the trained model, each feature is given a score to evaluate its feature importance \textit{based on the model}. We apply the \ac{ERFC}, given $N$ data points, introduced in \cite{johnsen_new_2021}, which is basically a summary score from the corresponding \ac{SHAP} values for each feature and individual data point,
\begin{equation}
    \kappa_k = \sum_{i=1}^N \frac{|\phi^{\SHAP}_{i,k}(\textbf{x}_i,\hat{y})|}{|\phi^{\SHAP}_0| + \sum_{j=1}^K |\phi^{\SHAP}_{i,j}(\textbf{x}_i,\hat{y})|} \,,
\label{ERFC}
\end{equation}
with $\phi_0^{\SHAP} = v_{\textbf{x},\hat{y}}(\emptyset) $. The \acp{ERFC} scores can be computed based on the data used to construct the model, as we only need to measure what the model considers important. The features with the largest \ac{ERFC}-values are then considered the most promising ones \textit{based on the model}. Depending on your hypothesis of interest, one can evaluate the uncertainty in the feature importance by computing sub-\ac{SAGE} estimates with corresponding bootstrap-derived percentile intervals. However, it is important that the sub-\ac{SAGE} estimates are calculated based on independent test data never used during training. From the trained model, we compute the \ac{ERFC} based on the training data and validation data together ($n=12800$), and~\cref{RankTable} shows the top 10 features with the largest \ac{ERFC}-values.
\begin{table}
\centering
    \caption{\label{RankTable}The resulting ranking based on the \acf{ERFC} after having trained an XGBoost model consisting of 6 influential features and 94 noise features.}
    \begin{tabular}{ll}
    \toprule
    \textbf{Feature} & \textbf{ERFC}  \\ \midrule
    $x_6$ & 0.48 \\ \midrule
    $x_5$ & 0.060 \\ \midrule
    $x_3$ & 0.026 \\ \midrule
    $x_1$ & 0.022 \\ \midrule
    $x_4$ & 0.0036 \\ \midrule
    $x_2$ & 0.0030 \\ \midrule
    $x_{12}$ & 0.0028  \\ \midrule
    $x_{30}$ & 0.0022 \\ \midrule
    $x_{40}$ & 0.0019\\ \bottomrule
    \end{tabular}
\end{table}
This shows that the XGBoost model has accurately ranked the most influential features among the top $10$ list, for this rather simple relationship. These scores, based on \ac{SHAP} values, are only with respect to what the \textit{model} considers important. The sub-\ac{SAGE} can now be applied to infer whether the importance of any feature from the model is also reflected in the data. As an example, let us consider features $6$, $1$, $2$ and $12$ where feature $6$ has a strong influence, feature $1$ has a weaker influence, and feature $2$ has the weakest influence, while feature $12$ has no influence with respect to $f(\textbf{x}_i)$ in~\cref{SimRegEx}. Their sub-\ac{SAGE} estimate along with histograms to estimate the corresponding distribution of the sub-SAGE estimators are shown in~\cref{Boothists} for training plus validation data as well as for independent test data. We see that sub-\ac{SAGE} values inferred using training data overestimates the false influence of feature $12$, while using the test data correctly indicates that feature $12$ has a weak or no influence. We also see from the other histograms that using the training data underestimates the uncertainty in the sub-SAGE estimate. 

By using the test data for computation of the sub-\ac{SAGE} estimates, the estimated $95\%$ percentile intervals of the sub-\ac{SAGE} values for each feature are $6: (39.45,~44.15)$, $1: (-0.038,~0.14)$, $2: (-0.043,~0.040)$ and $12: (-0.030,~0.0050)$. These ranges allow us to conclude that feature $6$, correctly, is highly influential, while feature $12$ is highly unlikely to have any influence. The added benefit of the estimated confidence intervals is to prevent us from concluding that features $1$ and $2$ are influential but rather concluding that feature $1$ is highly likely to be influential, as its average is above zero.

\begin{figure}[b!]
    \centering
    \includegraphics[width=0.97\textwidth]{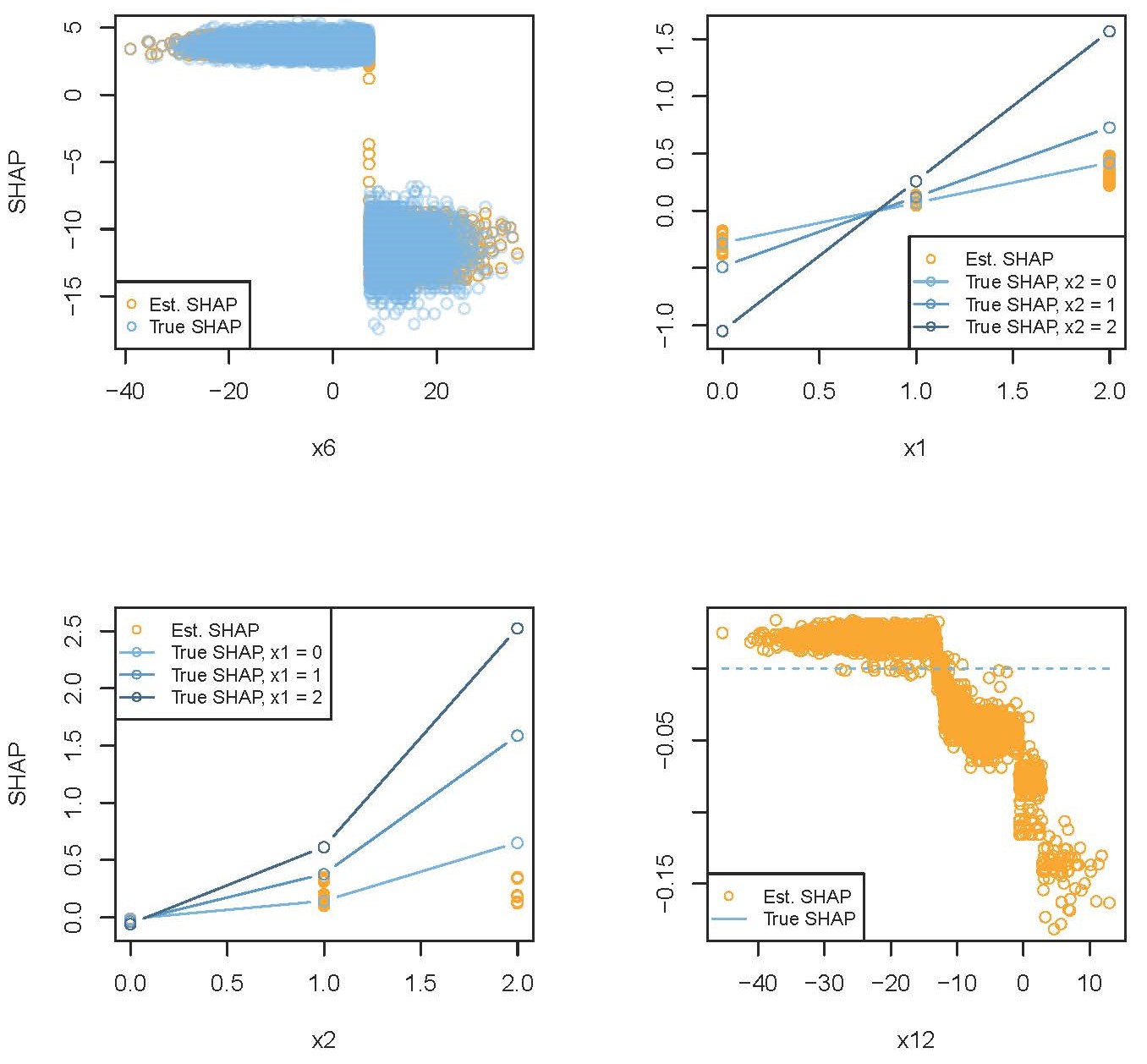}
    \caption{\label{estshapvStrue} Comparison of true \ac{SHAP} value for each data point with the estimated \ac{SHAP} value from the model fitted on the synthetic data, \cref{SimRegEx}. The deviations explain the reasons behind under- and overestimation of feature importance.}
\end{figure}

To correct for a potential bias in the plug-in estimator of the sub-SAGE as well as potential changes in the standard deviation of the estimator at different levels, the bias-corrected and accelerated bootstrap confidence interval may give more accurate bootstrap confidence intervals \citep{efron_better_1987}. This results in the following intervals $6:(39.45,~44.13)$, $1:(-0.034,~0.14)$, $2:(-0.047,~0.037)$ and $12:(-0.031,~0.0040)$, with only negligible changes from the percentile confidence intervals. The sub-SAGE underestimation of the influence of both features $1$ and $2$, but particularly feature $2$, can be explained by looking at~\cref{estshapvStrue}.
\FloatBarrier
As the data generating process is known, we can compare the true \ac{SHAP} value at each point with the corresponding \ac{SHAP} value from the fitted model. 
It shows that the influence of feature $6$ is quite accurately modelled, while the effect of feature $1$ and particularly feature $2$ is highly underestimated when $x_1 = 1$ and $x_2 = 2$. As features 1 and 2 interact, the \ac{SHAP} value of feature $1$ depends on the value of feature $2$. It also becomes clear that feature $12$, according to the model, has a negative trend in the \ac{SHAP} value, but the true \ac{SHAP} value is equal to zero (no importance), regardless of the value of feature $12$. See \ref{shapcomps} for derivations. 

\section{Application on genetic data using the UK Biobank resource} \label{sec:genetic}
To demonstrate the ability of sub-SAGE on observed data, we consider a realistic high-dimensional machine learning problem that often occurs when using genetic data, namely the influence of specific features on a given trait.

We use both genetic and non-genetic data from UK Biobank, a large prospective cohort study in the United Kingdom that began in 2006 consisting of about $500'000$ participants~\citep{sudlow2015uk,bycroft2018uk}, and attempt to infer the influence of specific features with respect to obesity ($\BMI \geq 30$), by training an XGBoost model and computing sub-SAGE values. 

We treat this as a classification problem between the categories obese and non-obese~\citep[see][for details]{johnsen_new_2021}. Of particular interest is whether any genetic markers are important. The most used method in this setting is a so-called \ac{GWAS}, where each genetic variant is tested individually in a general linear (mixed-effects) regression model~\citep{visscher_10_2017,zhou_efficiently_2018}. A corresponding $p$-value less than $5\times 10^{-8}$ is often considered statistically significant, a tiny significance level due to the multiple comparison problem~\citep{goeman_multiple_2014}. When the same association is replicated in an independent data set, the association is considered to be robust.

We study the XGBoost model constructed in~\citet{johnsen_new_2021} based on $3000$ features both genetic (\ac{SNP}) and non-genetic, for $64'000$ unrelated White-British participants from UK Biobank. The genetic data consists of so-called \textit{minor allele counts} or genotype values from \acp{SNP}~\citep[see e.g.][]{visscher_10_2017} filtered to ensure independence without significant loss of information~\citep{johnsen_new_2021}. Non-genetic features included are sex, age, physical activity frequency, intake of saturated fate, sleep duration, stress and alcohol consumption~\citep[see][for definitions]{johnsen_new_2021}. The model is trained with hyperparameters: learning rate $\eta = 0.05$, $colsample = subsample = colsample\_by\_tree = 0.8$, $max\_depth=2$, $\lambda=1$, $\gamma=1$, $early\_stopping\_rounds = 20$, and binary cross-entropy loss. The trained model included only $532$ features among the $3000$ input features spread along a total of $607$ trees. The features with the largest \ac{ERFC}-scores, based on the training data, and therefore considered the most promising features, are given in \cref{RankTableObesity}.
\begin{table}[t]
\centering
    \caption{The resulting ranking based on the \acf{ERFC} after having trained an XGBoost model consisting of $3000$ features and $64'000$ individuals from UK Biobank.}
    \begin{tabular}{ll}
    \toprule
    \textbf{Feature} & \textbf{ERFC}  \\ \midrule
    Alcohol intake frequency & 0.088 \\ \midrule
    Genetic sex & 0.086 \\ \midrule
    Physical activity frequency & 0.073 \\ \midrule
    Intake of saturated fat & 0.044 \\ \midrule
    Sleep duration &  0.036 \\ \midrule
    Stress & 0.034 \\ \midrule
    Age at recruitment & 0.033  \\ \midrule
    rs17817449 & 0.017 \\ \midrule
    rs489693 & 0.012\\ \midrule
    rs1488830 & 0.011 \\ \midrule 
    rs13393304 & 0.010 \\ \midrule 
    rs10913469 &  0.01 \\ \midrule
    rs2820312 & 0.0086 \\ \bottomrule
    \end{tabular}
    \label{RankTableObesity}
\end{table}

\begin{figure}[t!]
    \centering
    \includegraphics[width=0.8\textwidth]{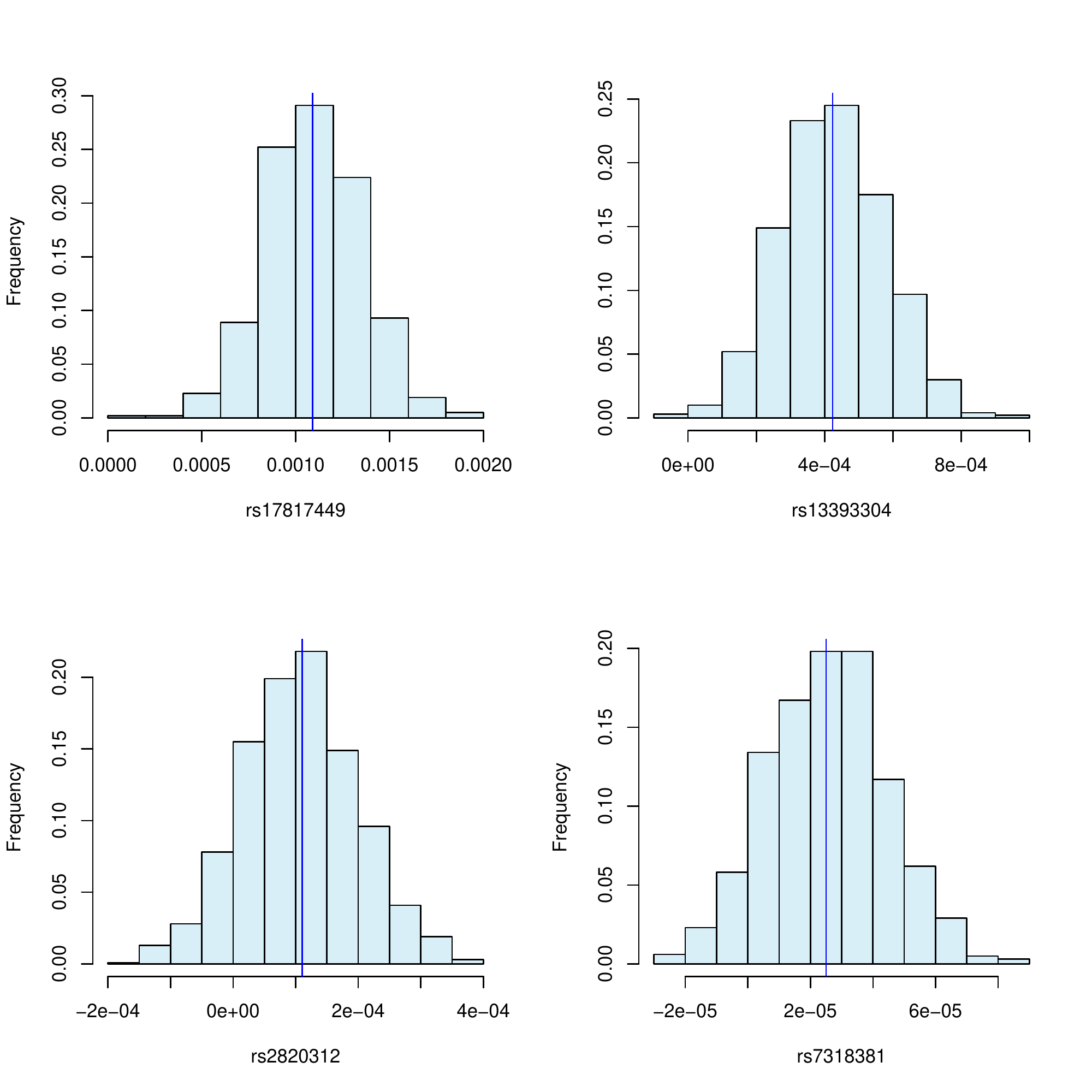}
    \caption{The estimates and corresponding uncertainties in the sub-SAGE values for the four \acp{SNP} agree with previous studies (\ac{GWAS}) regarding \ac{SNP}-association with obesity. }
\label{subsageObesity}
\end{figure}

While the non-genetic features are considered the most important, the most important \ac{SNP} according to the model is rs17817449, a \ac{SNP} connected to the FTO gene at chromosome 16, previously associated (statistically significant) with obesity in a large number of genome-wide association studies including different independent data sets~\citep{locke_genetic_2015}. The \ac{SNP} rs13393304 at chromosome 2 has previously been associated with obesity using UK Biobank data~\citep{karlsson2019contribution}. The \ac{SNP} rs2820312 has not previously been associated with obesity, but with hypertension based on UK Biobank data~\citep{gagliano_taliun_exploring_2020}. The \acp{SNP} mentioned above are explored further by computing sub-SAGE estimates including paired bootstrap-derived percentile intervals by using $20'000$ (unrelated White-British) participants from UK Biobank not used while training the model. We also compute sub-SAGE for the randomly selected \ac{SNP} rs7318381, which has never been associated with obesity, and with a small \acp{ERFC} in the XGBoost model $(0.0016)$. The result is given in~\cref{subsageObesity}. 

The sub-SAGE values do indicate that both rs17817449 and rs13393304 are highly likely to be associated with obesity. The $95 \%$ percentile interval of the sub-SAGE value for rs17817449 is $(0.0006,~0.0016)$, and $(0.00014,~0.00073)$ for rs13393304. The \acp{SNP} rs2820312 and rs7318381 are less likely to be associated with obesity, and if they are true associations, the uncertainties in the estimates indicate that the effects are microscopic. The $95\%$ percentile intervals for rs2820312 is $(-7.08\cdot 10^{-5},~2.95\cdot 10^{-4})$, and $(-1.13\cdot 10^{-5},~6.32\cdot 10^{-5})$ for rs7318381.

When dealing with relatively large data sizes such as for the genetic example above, the bias-corrected and accelerated bootstrap interval can become infeasible due to the estimation of the acceleration parameter. However, as the acceleration parameter is proportional to the skewness of the bootstrap distribution, and if the bootstrap distribution indeed has a small skewness, as is the case here, it is often sufficient to set the acceleration parameter equal to zero. This gives no change in the percentile intervals of rs17817449 and rs13393304, but the bias-corrected $95\%$ bootstrap intervals of rs2820312 and rs7318381 become $(-6.10\cdot 10^{-5},~0.00030)$ and $(-1.18\cdot 10^{-5},~6.19\cdot 10^{-5})$ respectively. These are negligible changes, indicating that the plug-in estimates are low-biased.

\FloatBarrier
\section{Discussion and conclusion} \label{sec:discussion}

We present a Shapley value based framework for inferring the importance of individual features, including uncertainty in the estimator. We demonstrate how to infer feature influence for a tree ensemble model with high-dimensional data using sub-SAGE and paired bootstrapping. As an example, we use XGBoost, a gradient tree-boosting model, applied to both a known data generating process, as well as realistic high-dimensional data. We emphasize the importance of using test data, independent of data used to construct the model, to compute sub-SAGE estimates.

It is important to notice that the percentile intervals, constructed to evaluate the uncertainty in the sub-SAGE estimate, themselves include uncertainty. The uncertainty of the percentile intervals depends on the number of bootstraps, $B$, as well as the size $n$ of data. However, in addition, the uncertainty also depends on the ratio $p/n$, where $p$ is the total number of features \textit{used} in the model (not necessarily the number of input-features for constructing the model). This fact is particularly important in high-dimensional problems, and it has been discussed for instance in~\cite{boothighdim}. When applied to linear models, one observation from a simulation is for instance that the paired bootstrap becomes more conservative (loss of power) the larger the ratio $p/n$ is. Observe that for the simulation example above, $p/n = 62/3200 = 0.019$, while for the genetic data, the ratio is $p/n = 533/20000 = 0.027$, deliberately chosen to be small in order to account for the problems arising when $p/n$ becomes too large. For the genetic data, a filtering process is first needed as the data from UK Biobank originally includes around $530'000$ \acp{SNP} and $207'000$ individuals $(p/n=2.56)$. 
The applied filtering method and potential pitfalls are described in~\cite{johnsen_new_2021}.

It seems reasonable to apply the same loss function in the sub-SAGE estimate as the loss function that was used to construct the model. However, there may be situations where it is meaningful to compute the sub-\ac{SAGE} values for a different loss function than the loss function used during training in order to make more objective interpretations. This may e.g. be the case when the model is provided `as is' and you do not know the training loss function, or when using adapted loss functions, e.g. weighted binary cross-entropy, but the interpretation is relevant for a standard cross-entropy.

In this work we have assumed all features to be mutually statistically independent, an unrealistic scenario in most cases, except for situations such as with genetic data where one can make sure that the genetic distance between the \acp{SNP} is sufficiently large to minimize the correlation. If many features are statistically dependent, one is required to estimate conditional expected values~\citep[see e.g.][]{aas_explaining_2021}. In a high-dimensional setting, this often becomes very tedious and even infeasible in most cases. 
An important line of future research to allow for easy evaluation of feature influence in a high-dimensional setting, is dimensionality reduction of the features with reduced loss of interpretation of the cluster variables created. 

\section{Acknowledgements}
This research was supported by the Norwegian Research Council grant 272402 (PhD Scholarships at SINTEF), project number 304843 (the EXAIGON project), as well the funding for research stays abroad for doctoral and postdoctoral fellows financed by the Norwegian Research Council. The research has been conducted using the UK Biobank Resource under Application Number 32285. We thank the Yale Center for Research Computing for guidance and use of the research computing infrastructure. We thank The Gemini Center for Sepsis Research for establishing cooperation with Yale School of Public Health.

\section{Code availability}
Source code is available at \url{https://github.com/palVJ/subSAGE}. 

\appendix
\section{The weights in sub-SAGE}\label{app:subsage_weights}
The sub-SAGE, $\psi_k$, is defined as~\cref{RedSAGEDef} and repeated here for convenience,
\begin{equation}
\psi_{k}(\textbf{X},Y,\hat{y}) = \sum_{\mathcal{S} \in \mathcal{Q}_k} \frac{|\mathcal{S}|!(M-|\mathcal{S}|-1)!}{3(M-1)!}\left[w_{\textbf{X},Y,\hat{y}}(\mathcal{S} \cup \{k\})-w_{\textbf{X},Y,\hat{y}}(\mathcal{S})\right] \,,
\end{equation}
with $\mathcal{Q}_k$ consisting of the subsets $\{\emptyset\}$, $\{m\}$ for $m = 1,\ldots,k-1,k+1,\ldots,M$ and $\{1,2,\ldots,k-1,k+1,\ldots,M\}$. In other words, there are three different achievable subset sizes, namely of size zero, one and $M-1$. As we want the sum of all weights to be equal to one, and that the sum of the weights of equal subset size is the same for all subset sizes, we need the corresponding weight for $\s = \{\phi\}$ and $\s = \{1,2,\ldots,k-1,k+1,\ldots,M\}$ to be $1/3$, while the sum of the weights for $\s = \{m\}$ for $m = 1,\ldots,k-1,k+1,\ldots,M$ needs to be $1/3$. For $\s = \{\phi\}$, we see that the weight is $0!(M-1)!/3(M-1)! = 1/3$ and for $\s = \{1,2,\ldots,k-1,k+1,\ldots,M\}$ the weight is $(M-1)!0!/3(M-1)! = 1/3$, just as we wanted. For the subsets of size one, the weight is $1!(M-2)!/3(M-1)! = 1/3(M-1)$. There are $M-1$ subsets of size one in total, and so the sum of the weights are also $1/3$. In other words, the definition of the weights in sub-SAGE makes sure that the sum of all weights is equal to one, and that the sum of the weights of equal subset size is the same for all subset sizes.

\section{Derivation of Sub-SAGE for squared error and binary cross-entropy}
\label{app:SAGE_trees}
Using as loss function the squared error loss, the loss per sample is $\ell = (y-\hat{y})^2$. Considering a feature $k$ for which to compute the sub-SAGE value, we separate the trees in our ensemble model into two groups: $\tau_{k}$, being the set of trees including feature $k$ as a splitting point, and its complement group ($\bar{\tau}_k$). Then, for any $\s \in \mathcal{Q}_k$,
\begingroup\makeatletter\def\f@size{8}\check@mathfonts
\begin{align}
\begin{split}
    &w_{\textbf{X},Y,\hat{y}}(\mathcal{S} \cup \{k\})-w_{\textbf{X},Y,\hat{y}}(\mathcal{S})  \\
    &=  E_{\mathbf{X},Y} \left[\left(Y\left(\textbf{X}\right)-V_{\textbf{X},\hat{y}}\left(\mathcal{S}\right)\right)^2\right] - E_{\mathbf{X},Y}[{\left(Y\left(\textbf{X}\right)-V_{\textbf{X},\hat{y}}(\mathcal{S} \cup \{k\})\right)}^2]  \\
        &= E_{\mathbf{X},Y}\left[\left(Y-\sum_{j \in \tau_k} V_{\textbf{X},f_j}\left(\mathcal{S}\right) 
        - \sum_{j \notin \tau_k} V_{\textbf{X},f_j}\left(\mathcal{S}\right) \right)^2
        - \left(Y-\sum_{j \in \tau_k} V_{\textbf{X},f_j}(\mathcal{S} \cup \{k\})
        - \sum_{j \notin \tau_k} V_{\textbf{X},f_j}(\mathcal{S} \cup \{k\})\right)^2\right] \\
        &=  E_{\mathbf{X},Y}\left[\left(Y-\sum_{j \in \tau_k} V_{\textbf{X},f_j}\left(\mathcal{S}\right) 
        - \sum_{j \notin \tau_k} V_{\textbf{X},f_j}\left(\mathcal{S}\right) \right)^2
        - \left(Y-\sum_{j \in \tau_k} V_{\textbf{X},f_j}(\mathcal{S} \cup \{k\})
        - \sum_{j \notin \tau_k}          V_{\textbf{X},f_j}\left(\mathcal{S}\right)\right)^2\right] \\
        &=  E_{\mathbf{X},Y}\left[
        2Y \left(\sum_{j \in \tau_k} V_{\textbf{X},f_j}(\mathcal{S} \cup     \{k\})- V_{\textbf{X},f_j}\left(\mathcal{S}\right)\right) 
        + \left(\sum_{j \in \tau_k}     V_{\textbf{X},f_j}\left(\mathcal{S}\right)\right)^2 
        - \left(\sum_{j \in \tau_k} V_{\textbf{X},f_j}(\mathcal{S} \cup     \{k\})\right)^2 \right. \\
        &+ \left.2\left(\sum_{j \notin \tau_k}     V_{\textbf{X},f_j}\left(\mathcal{S}\right)\right)
        \left(\sum_{j \in \tau_k} V_{\textbf{X},f_j}(\mathcal{S} \cup \{k\})-     V_{\textbf{X},f_j}\left(\mathcal{S}\right)\right)\right] \,, 
    \end{split}
    \label{SAGE_trees_full}
\end{align}
\endgroup
having used that the two random variables $V_{\textbf{X},f_j}(\mathcal{S} \cup \{k\})$ and $V_{\textbf{X},f_j}\left(\mathcal{S}\right)$ are equivalent, or equal in distribution, for $j \notin \tau_k$. Note that the corresponding observed value $v_{\textbf{x},f_j}(\suk) = E_{\textbf{X}_{\overline{\s}}}[f_j(\textbf{X}|\textbf{X}_{\s} = \textbf{x}_{\suk})] = E_{\textbf{X}_{\overline{\s}}}[f_j(\textbf{X}|\textbf{X}_{\s} = \textbf{x}_{\s})] = v_{\textbf{x},f_j}(\s)$ for all $\s \in \mathcal{Q}_k$ since the regression tree $f_j$ does not include feature $k$, and the features are assumed mutually independent.

Using as loss function the binary cross-entropy, the loss function per sample is $\ell = -y \log \hat{y} -(1-y)\log(1-\hat{y}) = (1-y)\sum_{\tau=1}^T f_{\tau} + \log \left(1+e^{-\sum_{\tau=1}^T f_{\tau}}\right)$. Then, we have
\begingroup\makeatletter\def\f@size{8}\check@mathfonts
\begin{align}
    \begin{split}
    &w_{\textbf{X},Y,\hat{y}}(\mathcal{S} \cup \{k\})-w_{\textbf{X},Y,\hat{y}}(\mathcal{S}) \\
    &=E_{\textbf{X},Y} \left[\left(1-Y\left(\textbf{X}\right)\right)\sum_{\tau=1}^{T} V_{\textbf{X},f_{\tau}}\left(\mathcal{S}\right) + \log \left(1+\exp \left(-\sum_{\tau=1}^{T} V_{\textbf{X},f_{\tau}}\left(\mathcal{S}\right)\right)\right) \right] \\
    &-E_{\textbf{X},Y}\left[\left(1-Y\left(\textbf{X}\right) \right)\sum_{\tau=1}^T V_{\textbf{X},f_{\tau}}(\mathcal{S} \cup \{k\}) + \log \left(1+\exp \left(-\sum_{\tau=1}^T V_{\textbf{X},f_{\tau}}(\mathcal{S} \cup \{k\})\right)\right)\right] \\
    &= E_{\textbf{X},Y}\left[\left(1-Y\left(\textbf{X}\right)\right) \left(\sum_{j \in \tau_k} V_{\textbf{X},f_j}(\mathcal{S}) + \sum_{j \notin \tau_k} V_{\textbf{X},f_j}\left(\mathcal{S}\right)\right) \right] \\
    &+ E_{\textbf{X},Y} \left[ \log \left(1+\exp \left(-\sum_{j \in \tau_k} V_{\textbf{X},f_j}\left(\mathcal{S}\right) - \sum_{j \notin \tau_k} V_{\textbf{X},f_j}\left(\mathcal{S}\right) \right) \right) \right] \\
    &- E_{\textbf{X},Y}\left[\left(1-Y \left(\textbf{X}\right)\right) \left(\sum_{j \in \tau_k} V_{\textbf{X},f_j}(\mathcal{S} \cup \{k\}) + \sum_{j \notin \tau_k} V_{\textbf{X},f_j}(\mathcal{S} \cup \{k\})\right) \right] \\
    &-E_{\textbf{X},Y}\left[ \log  \left(1+\exp \left(-\sum_{j \in \tau_k} V_{\textbf{X},f_j}(\mathcal{S} \cup \{k\}) - \sum_{j \notin \tau_k} V_{\textbf{X},f_j}(\mathcal{S} \cup \{k\}) \right) \right) \right] \\
    &= E_{\textbf{X},Y} \left[ \left(1-Y \left(\textbf{X}\right)\right) \left(\sum_{j \in \tau_{k}} V_{\textbf{X},f_j}\left(\mathcal{S}\right) - V_{\textbf{X},f_j}(\mathcal{S} \cup \{k\}) \right) \right. \\
    &\left. \quad + \log \left( \frac{1+\exp \left(-\sum_{j \in \tau_k} V_{\textbf{X},f_j}\left(\mathcal{S}\right) - \sum_{j \notin \tau_k} V_{\textbf{X},f_j}\left(\mathcal{S}\right) \right)}{1+\exp \left(-\sum_{j \in \tau_k} V_{\textbf{X},f_j}(\mathcal{S} \cup \{k\}) - \sum_{j \notin \tau} V_{\textbf{X},f_j}(\mathcal{S} \cup \{k\}) \right)} \right) \right] \,.
    \end{split}
\end{align}
\endgroup

\section{(Sub-)SAGE with multiple linear regression}\label{linregApp}
Consider a fitted linear regression model $\hat{y}_i =  \hat{\bm \beta}^{T} \textbf{x}_i$, with uncorrelated features. By applying the squared error loss, and by considering $\hat{\bm \beta}$ as a constant (by using data not used to estimate $\hat{\bm \beta}$), we have for a feature $k$, and a subset $\s \in \mathcal{Q}_k$ that
\begingroup\makeatletter\def\f@size{10}\check@mathfonts
\begin{align}
    \begin{split}
    &w_{\textbf{X},Y,\hat{y}}(\suk)-w_{\textbf{X},Y,\hat{y}}(\s) = E_{\bm X,Y}[(Y-V_{\bm X,\hat{y}}(\s))^2]-E_{\bm X,Y}[(Y-V_{\bm X,\hat{y}}(\suk))^2]\\
    & = E_{\textbf{X},Y}[2 Y\hat{\beta}_k(X_{k}-E[X_{k}]) + V_{\bm X,\hat{y}}(\s)^2-V_{\bm X,\hat{y}}(\suk)^2] \\
    & = 2\hat{\beta}_kE_{\textbf{X},Y}\left[Y(X_{k}-E[X_{k}])\right] + 2E_{\textbf{X},Y}
    \left[\hat{\beta}_k(\hat{\beta}^{T}_{S}X_{S} + \hat{\beta}^{T}_{\overline{\suk}}X_{\overline{\suk}}])(E[X_{k}]-X_{k})\right] \\
    &- \hat{\beta}^2_{k}E_{\textbf{X},Y}\left[\left({X_{k}}^2-{E[X_{k}]}^2 \right) \right]\\
    &= 2\hat{\beta}_k \Cov(Y,X_k) - \hat{\beta}^2_{k} \Var(X_k),
    \end{split}
\label{LinRegEx}
\end{align}
\endgroup
with
\[V_{\bm X,\hat{y}}(\s) = \hat{\beta}_k E[X_k] + \hat{\beta}_{\s}X_s + \hat{\beta}_{\overline{\suk}} E[X_{\overline{\suk}}],\] 
the stochastic version of $v_{\bm x,\hat{y}}(\s) = E[\hat{y}(\bm X) | X_{\s}= x_{\s}] = \hat{\beta}_k E[X_k] + \hat{\beta}_{\s}x_s + \hat{\beta}_{\overline{\suk}} E[X_{\overline{\suk}}]$, and
\[V_{\bm X,\hat{y}}(\suk) = \hat{\beta}_k X_k + \hat{\beta}_{\s}X_s + \hat{\beta}_{\overline{\suk}} E[X_{\overline{\suk}}],\]

the stochastic version of $v_{\bm x,\hat{y}}(\suk)$. See Appendix B in~\cite{aas_explaining_2021} for proof of $v_{\bm x,\hat{y}}(\s)$ in linear regression. The second term in the third line of \cref{LinRegEx} is equal to zero since the features are independent, and $\hat{\bm \beta}$ is considered a constant. Notice therefore that the sub-SAGE value, as well as the \ac{SAGE}-value, is independent of the subset $\s$ used, and equal to \cref{LinRegEx}. 

The second term $\hat{\beta_k}^2 \Var(X_k)$ is in fact equal to the increased variance in the model by including feature $k$ actively in the model since 
\begin{align}
    \begin{split}
        &E[V_{\bm X,\hat{y}}(\s)^2-V_{\bm X,\hat{y}}(\suk)^2] \\
        &= E[V_{\bm X,\hat{y}}(\s)^2] - E[V_{\bm X,\hat{y}}(\s)]^2-(E[V_{\bm X,\hat{y}}(\suk)^2]-E[V_{\bm X,\hat{y}}(\suk)]^2) \\
        &= \Var(V_{\bm X,\hat{y}}(\s)) - \Var(V_{\bm X,\hat{y}}(\suk)),
    \end{split}
\end{align}
because $E[V_{\bm X,\hat{y}}(\s)] = E[V_{\bm X,\hat{y}}(\suk)]$.

For linear regression models, this shows that the sub-SAGE value is only positive if the agreement between the model and the independent test data (first term in \cref{LinRegEx}) upweights the increased variance in the model (second term in \cref{LinRegEx}) by including feature $k$. 

We neither know the variance of $X_k$ nor the correlation between $X_k$ and $Y$, and so these must also be estimated from the data. The sample mean and sample covariance are unbiased and consistent estimators. Therefore, by using \textit{independent} test data $(\textbf{x}^{0}_1,y^{0}_1),...,(\textbf{x}^{0}_{N_I},y^{0}_{N_I})$ of size $N_I$, the estimator of $\hat{\beta}_k$, denote it $T(\hat{\beta}_k$), is statistically independent from the test data, and by applying the sample mean and covariance we get the following unbiased estimate of~\cref{LinRegEx}
\begingroup\makeatletter\def\f@size{8}\check@mathfonts
\begin{align}
    \begin{split}
         & \hat{w}_{\textbf{X},Y,\hat{y}}(\suk)-\hat{w}_{\textbf{X},Y,\hat{y}}(\s) \\
        & = \frac{2\hat{\beta}_j}{n_I-1} \sum_{i=1}^{N_I} \left[y^{0}_i x^{0}_{i,j}-\left(\frac{1}{N_I}\sum_{i=1}^{N_I} x_{i,j}\right)\left(\frac{1}{N_I}\sum_{i=1}^{N_I} y_{i,j}\right) \right] - \hat{\beta_j}^2\frac{1}{N_I-1}\sum_{i=1}^{N_I} \left(x_{i,j}-\frac{1}{N_I}\sum_{i=1}^{N_I} x_{i,j}\right)^2 \\
         & = 2 \hat{\beta}_j \widehat{\Cov}^{0}(Y,X_k) - \hat{\beta_k}^2 \widehat{\Var}^{0}(X_k) \,.
    \end{split}
\label{EstLinRegEx}
\end{align}
\endgroup

If we did not use training data separately for constructing the model, and test data to compute sub-SAGE values, the second term in the third line of \cref{LinRegEx} would no longer become zero since the estimator $T(\hat{\bm \beta})$ naturally is correlated with the training data itself. It may seem confusing to treat $\hat{\beta}_k$ in \cref{LinRegEx} as a constant when the corresponding estimator $T(\hat{\beta}_k)$ indeed has a distribution based on the training data. However, one may look at the procedure of sub-SAGE as objectively observing the properties of the raw model itself without taking into account the data used for training the model. 

\section{Sub-SAGE estimate for tree ensemble models with tree stumps}\label{app:tree_stumps}
Consider a tree ensemble model with regression trees of depth one, so-called tree stumps. Each tree stump includes exactly one feature from the set $\mathcal{M}$ of all $M$ features. In accordance with earlier notation, let $\tau_k$ denote the set of tree stumps that include feature $k$. Then, \cref{SAGE_trees_full} reduces to
\begingroup\makeatletter\def\f@size{8}\check@mathfonts
\begin{align}
\begin{split}
&w_{\textbf{X},Y,\hat{y}}(\mathcal{S} \cup \{k\})-w_{\textbf{X},Y,\hat{y}}(\mathcal{S})  \\
&=  E_{\mathbf{X},Y}\left[
    2Y \left(\sum_{j \in \tau_k} V_{\textbf{X},f_j}(\mathcal{S} \cup \{k\})- V_{\textbf{X},f_j}\left(\mathcal{S}\right)\right) 
    + \left(\sum_{j \in \tau_k} V_{\textbf{X},f_j}\left(\mathcal{S}\right)\right)^2 
    - \left(\sum_{j \in \tau_k} V_{\textbf{X},f_j}(\mathcal{S} \cup \{k\})\right)^2 \right. \\
    &+ \left.2\left(\sum_{j \notin \tau_k}     V_{\textbf{X},f_j}\left(\mathcal{S}\right)\right)
    \left(\sum_{j \in \tau_k} V_{\textbf{X},f_j}(\mathcal{S} \cup \{k\})-     V_{\textbf{X},f_j}\left(\mathcal{S}\right)\right)\right] \\ 
    & = 2Cov\left(Y,\sum_{j \in \tau_k} f_j(X_{k})\right) - 
    Var \left(\sum_{j \in \tau_k} f_j(X_{k}) \right) \,,
\label{treestumps}   
\end{split}
\end{align}
\endgroup
because all random variables $V_{\textbf{X},f_j}(\s)$ for $j \notin \tau_k$, for every $\s$ are now independent of all $V_{\textbf{X},f_j}(\s)$ and $V_{\textbf{X},f_j}(\suk)$ for $j \in \tau_k$. Further, for every $j \in \tau_k$, $V_{\textbf{X},f_j}(\s) = E_{\textbf{X}}[f_j(\textbf{X})]$, a constant equal to the expected value of the output of the regression tree $f_j$, and $E_{\textbf{X}}[V_{\textbf{X},f_j}(\suk)] = E_{\textbf{X}}[f_j(\textbf{X})]$, since the regression tree $f_j$ only includes feature $k$. 
Therefore, the last term in~\cref{SAGE_trees_full} vanishes.  
Observe that, in the case of tree stumps,
\begin{align*}
\begin{split}
   &E_{\textbf{X},Y}\left[Y\left(\sum_{j \in \tau_k} V_{\textbf{X},f_j}(\mathcal{S} \cup \{k\})- V_{\textbf{X},f_j}\left(\mathcal{S}\right)\right)\right]\\ 
   &= E_{\textbf{X},Y}\left[Y\left(\sum_{j \in \tau_k} f_j(X_k)\right)\right]-E_{Y}[Y]E_{\textbf{X}}\left[\sum_{j \in \tau_k} f_j(X_k)\right]
   = \Cov\left(Y,\sum_{j \in \tau_k} f_j(X_k)\right)\,.
\end{split}
\end{align*}
Likewise,

\begin{align*}
\begin{split}
    &E_{\textbf{X},Y}\left[\left(\sum_{j \in \tau_k} V_{\textbf{X},f_j}(\mathcal{S} \cup \{k\})\right)^2-\left(\sum_{j \in \tau_k} V_{\textbf{X},f_j}\left(\mathcal{S}\right)\right)^2 \right]\\
    &= E_{\textbf{X}}\left[\left(\sum_{j \in \tau_k} f_j(X_k)\right)^2\right] - E_{\textbf{X}}\left[\sum_{j \in \tau_k} f_j(X_k)\right]^2 = \Var\left(\sum_{j \in \tau_k} f_j(X_k)\right).
\end{split}
\end{align*}

Hence, the expression given in \cref{treestumps} independent of the subset $\s$. The expression in \cref{treestumps} is therefore also equal to the sub-SAGE value, $\hat{\psi}_k$ (or \ac{SAGE} value). Both the covariance and the variance need to be must be estimated in practice. Given independent test data $(\textbf{x}^{0}_1,y^{0}_1), \dots, (\textbf{x}^{0}_{N_I},y^{0}_{N_I})$, an unbiased estimate is given by
\begin{equation}
    \begin{split}
        \hat{\psi}_k = \frac{1}{N_I^{0}-1}\sum_{i=1}^{N_I^{0}} &\left(y^{0}_i-\sum_{i=1}^{N_I} y^{0}_i\right)
        \left(\sum_{j \in \tau_k} f_j(x^{0}_{i,k})-\sum_{j \in \tau_k}
        v_{x^0_{i,k},f_j}(\emptyset)\right) \\
        &- \frac{1}{N_I^{0}-1}\sum_{i=1}^{N_I^{0}} \left({\sum_{j \in \tau_k} f_j(x^{0}_{i,k}) - \sum_{j \in \tau_k} v_{x^0_{i,k},f_j}(\emptyset)}\right)^2\,.
        \label{sageunbiased}   
    \end{split}
\end{equation}

\section{Sub-SAGE properties related to Shapley values}\label{app:sage_axiom_properties}

Symmetry, null player, linearity, monotonicity and efficiency are all properties of Shapley values. Below we investigate whether the same properties apply for sub-SAGE values.

\subsection{Symmetry}
Given two features $j$ and $k$ such that $v(\mathcal{S} \cup \{j\}) = v(\mathcal{S} \cup \{k\}) $ for all $\mathcal{S} \in \{\mathcal{Q}_j,\mathcal{Q}_k\}$ in which $\{j,k\} \notin \mathcal{S}$. Then their sub-SAGE values indeed are identical, $\psi_j = \psi_k$, and so the symmetry property follows by definition. This means in practice that two perfectly correlated features have equal sub-SAGE values.

\subsection{Dummy property (null player)}
Given a feature $k$ where $v(\mathcal{S} \cup \{k\}) = v(\mathcal{S})$ for all $\mathcal{S} \in \mathcal{Q}_k$. Then $\psi_k = 0$, and so the dummy property follows by definition. 

\subsection{Linearity}
Given two value functions $v(\mathcal{S})$ and $w(\mathcal{S})$, the sub-SAGE value of the sum of the value functions $v(\mathcal{S}) + w(\mathcal{S})$ is equal to the sum of the sub-SAGE for each value function,
\begin{equation}
    \psi_k(v + w) = \psi_k(v) + \psi_k(w) \, .
\end{equation}

\subsection{Monotonicity}
Consider two models $\hat{f}_1$ and $\hat{f}_2$ used to predict the same relationship $y = f(\textbf{x})$, for the same features $\mathbf{x}$. If for any feature $k$ we have $v_{\hat{f}_1}(\mathcal{S} \cup \{k\}) - v_{\hat{f}_1}(\mathcal{S}) \ge v_{\hat{f}_2}(\mathcal{S} \cup \{k\}) - v_{\hat{f}_2}(\mathcal{S})  $ for all $\mathcal{S} \in \mathcal{Q}_k$, then by definition, $\psi^{\hat{f}_1}_k \ge \psi^{\hat{f}_2}_k$, with $\psi^{\hat{f}_1}_k$ the sub-SAGE value of feature $k$ when applying model $\hat{f}_1$ and $\psi^{\hat{f}_2}_k$ the corresponding sub-SAGE value when applying model $\hat{f}_2$. This means that an adjustment of model $\hat{f}_2$ to $\hat{f}_1$ such that feature $k$'s importance increases also increases its sub-SAGE value. Therefore, the monotonicity property follows by definition.

\subsection{sub-SAGE does not share the efficiency property}
Consider the definition of the Shapley value, $\phi_k$, applied on a specific value function $v$:
\begin{equation}
    \phi_{k} = \sum_{\mathcal{S} \subseteq \mathcal{M} \setminus \{k\}} \frac{|\mathcal{S}|!(M-|\mathcal{S}|-1)!}{M!}\left[v(\mathcal{S} \cup \{k\})-v(\mathcal{S})\right] \,.
\end{equation}
The efficiency property for the Shapley value reads
\begin{equation}
\sum_{k=1}^{M} \phi_k = v(\mathcal{M}) - v(\emptyset) \,,
\end{equation}
for $M$ "players". This can be observed more easily by using instead the following formulation of the Shapley value
\begin{equation}
    \phi_{k} = \frac{1}{M!}\sum_{R} \left[v(s_k(R) \cup \{k\})-v(s_k(R))\right] \,,
    \label{shapleyDef2}
\end{equation}
where the sum is over all \textit{orderings} $R$ of the $M$ features, with a total of $M!$ orders. The function $s_k(R)$ maps a given ordering $R$ and a particular feature $k$ to the corresponding subset of features preceding feature $k$ in the specific ordering. For instance, for $\mathcal{M} = \{1,2,3\}$, one possible ordering is $R=(2,3,1)$ with $s_1(R) = (2,3)$. We then have
\begin{align}
    \begin{split}
        &\sum_{k=1}^M \phi_{k} = \sum_{k=1}^M \frac{1}{M!}\sum_{R} \left[v(s_k(R) \cup \{k\})-v(s_k(R))\right] \\
        &= \frac{1}{M!}\sum_{R}\sum_{k=1}^M\left[v(s_k(R) \cup \{k\})-v(s_k(R))\right] \\
        &= \frac{1}{M!}\sum_{R} \left(v(\mathcal{M}) - v(\emptyset)\right)\\
        &= \frac{1}{M!} M!\left(v(\mathcal{M}) - v(\emptyset)\right) = v(\mathcal{M}) - v(\emptyset),
    \end{split}
\end{align}
since for a specific ordering $R$ and feature $k$, in the sum $\sum_{k=1}^M\left[v(s_k(R) \cup \{k\})-v(s_k(R))\right]$ all terms cancel each other, except $v(\mathcal{M})$ and $v(\emptyset)$.\\

The sub-SAGE value, $\psi_k$, for a feature $k$ is not a sum over all subsets $\mathcal{S} \subseteq \mathcal{M} \setminus \{k\}$, but limited to the sets in $\mathcal{Q}_k$,
\begin{equation}
    \psi_{k}(\textbf{y},\hat{\textbf{y}}) = \sum_{\mathcal{S} \in \mathcal{Q}_k} \frac{|\mathcal{S}|!(M-|\mathcal{S}|-1)!}{3(M-1)!}\left[{v\left(\s \cup \{k\}\right)-v\left(\s\right)}\right] \,,
\end{equation}
and therefore, from the definition in~\cref{shapleyDef2}, is \textit{not} the sum over all orderings $R$. The sub-SAGE value therefore does not share the efficiency property of the Shapley value.

\section{SHAP computations for fig. 3}\label{shapcomps}

Consider this time the SHAP value of a given data generating process, $f$, with known relationship:

\begin{equation}
    \phi^{\SHAP}_{k}(\textbf{x},f) = \sum_{\mathcal{S} \subseteq \mathcal{M} \setminus \{k\}} \frac{|\mathcal{S}|!(M-|\mathcal{S}|-1)!}{M!}\left[v_{\textbf{x},f}(\mathcal{S} \cup \{k\})-v_{\textbf{x},f}(\mathcal{S})\right],
\end{equation}

By applying the data generating process, $f$, explained in Section 5, the exact SHAP value of feature $1$ can be computed by partitioning in the subsets $\s$ \textit{not including} feature $2$, as well as those \textit{including} feature $2$. For all $\s$ not including feature $2$, and by using the result in Appendix B in \cite{aas_explaining_2021}:

\begin{align*}
 v_{\textbf{x}_i,f}(\mathcal{S} \cup \{k\})-v_{\textbf{x}_i,f}(\mathcal{S}) =  a_1 (x_{i,1}-E[X_1]) + a_{21}E[e^{X_2}](x_{i,1}-E[X_1]),
\end{align*}
 
independent of the subset $\s$ used. Of all $\s \subseteq \mathcal{M} \setminus \{1\}$, a half of them will not include feature $2$, and the sum of the corresponding Shapley weights are given by:

\begin{align*}
\begin{split}
 &\sum_{|\s|=0}^{M-2} \frac{|\s|!(M-|\s|-1}{M!}\binom{M-2}{|\s|} = \sum_{|\s|=0}^{M-2} \frac{|\s|!(M-|\s|-1}{M!}\frac{(M-2)!}{|\s|!(M-2-|\s|!)} \\
 &= \sum_{|\s|=0}^{M-2} \frac{1}{M} - \frac{1}{M(M-1)}\sum_{|\s|=0}^{M-2} |\s| = \frac{1}{2}.
\end{split}
\end{align*}
 
For all $\s$ including feature $2$:

\begin{align*}
 v_{\textbf{x}_i,f}(\mathcal{S} \cup \{k\})-v_{\textbf{x}_i,f}(\mathcal{S}) =  a_1 (x_{i,1}-E[X_1]) + a_{21}e^{x_{i,2}}(x_{i,1}-E[X_1]).
\end{align*}

As the sum of the Shapley weights are equal to one, the sum of the Shapley weights for these $\s$ must also be 1/2. Hence, the SHAP value of feature $1$ is given by:

\begin{align}
\begin{split}
&\phi_{i,1}(\textbf{x}_i) = \frac{1}{2}(a_1 (x_{i,1}-E[X_{1}]) + a_{21}E[e^{X_{2}}] (x_{i,1}-E[X_{1}])) \\
&+\frac{1}{2}(a_1(x_{i,1}-E[X_{1}])+a_{21} e^{x_{i,2}}(x_{i,1}-E[X_{1}]))\\
&= a_1(x_{i,1}-E[X_{1}])+ a_{21}E[e^{X_{2}}](x_{i,1}-E[X_{1}]) \\
&+ \frac{1}{2}a_{21}x_{i,1}(e^{x_{i,2}}-E[e^{X_{2}}])-\frac{1}{2}a_{21}E[X_{1}](e^{x_{i,2}}-E[e^{X_{2}}]).
\end{split}
\end{align}

In the exact same manner one can show that:

\begin{align}
\begin{split}
&\phi_{i,2}(\textbf{x}_i) = \frac{1}{2}(a_2(x_{i,2}-E[X_{2}])+a_{21}E[X_{1}](e^{x_{i,2}}-E[e^{X_{2}}])\\
&+ \frac{1}{2}(a_2(x_{i,2}-E[X_{2}])+a_{21}x_{i,1}(e^{x_{i,2}}-E[e^{X_{2}}])) \\
&= a_2(x_{i,2}-E[X_{2}]) + a_{21}E[X_{1}](e^{x_{i,2}}-E[e^{X_{2}}])\\
&+\frac{1}{2}a_{21}x_{i,1}(e^{x_{i,2}}-E[e^{X_{2}}])-\frac{1}{2}a_{21}E[X_{1}](e^{x_{i,2}}-E[e^{X_{2}}]).
\end{split}
\end{align}

\begin{align}
\begin{split}
&\phi_{i,6} = \frac{1}{2}(a_6 E[X_{5}](I(x_{i,6}>7)-E[I(X_{6}>7)]) \\
& \frac{1}{2}(a_6 x_{i,5}(I(x_{i,6}>7)-E[I(X_{6}>7)])) \\
&= a_6 E[X_{5}](I(x_{i,6}>7)-E[I(X_{6}>7)])\\
&+\frac{1}{2}a_6 I(x_{i,6}>7) (x_{i,5}-E[X_{5}])-\frac{1}{2}a_6 E[I(X_{6}>7)](x_{i,5}-E[X_{5}]).
\end{split}
\end{align}

\bibliographystyle{elsarticle-harv} 
\bibliography{references}

\begin{thebibliography}{32}
\expandafter\ifx\csname natexlab\endcsname\relax\def\natexlab#1{#1}\fi
\providecommand{\url}[1]{\texttt{#1}}
\providecommand{\href}[2]{#2}
\providecommand{\path}[1]{#1}
\providecommand{\DOIprefix}{doi:}
\providecommand{\ArXivprefix}{arXiv:}
\providecommand{\URLprefix}{URL: }
\providecommand{\Pubmedprefix}{pmid:}
\providecommand{\doi}[1]{\href{http://dx.doi.org/#1}{\path{#1}}}
\providecommand{\Pubmed}[1]{\href{pmid:#1}{\path{#1}}}
\providecommand{\bibinfo}[2]{#2}
\ifx\xfnm\relax \def\xfnm[#1]{\unskip,\space#1}\fi
\bibitem[{Aas et~al.(2021)Aas, Jullum and Løland}]{aas_explaining_2021}
\bibinfo{author}{Aas, K.}, \bibinfo{author}{Jullum, M.},
  \bibinfo{author}{Løland, A.}, \bibinfo{year}{2021}.
\newblock \bibinfo{title}{Explaining individual predictions when features are
  dependent: {More} accurate approximations to {Shapley} values}.
\newblock \bibinfo{journal}{Artificial Intelligence} \bibinfo{volume}{298}.
\bibitem[{Bycroft et~al.(2018)Bycroft, Freeman, Petkova, Band, Elliott, Sharp,
  Motyer, Vukcevic, Delaneau, O’Connell et~al.}]{bycroft2018uk}
\bibinfo{author}{Bycroft, C.}, \bibinfo{author}{Freeman, C.},
  \bibinfo{author}{Petkova, D.}, \bibinfo{author}{Band, G.},
  \bibinfo{author}{Elliott, L.T.}, \bibinfo{author}{Sharp, K.},
  \bibinfo{author}{Motyer, A.}, \bibinfo{author}{Vukcevic, D.},
  \bibinfo{author}{Delaneau, O.}, \bibinfo{author}{O’Connell, J.}, et~al.,
  \bibinfo{year}{2018}.
\newblock \bibinfo{title}{The uk biobank resource with deep phenotyping and
  genomic data}.
\newblock \bibinfo{journal}{Nature} \bibinfo{volume}{562},
  \bibinfo{pages}{203--209}.
\bibitem[{Chen and Guestrin(2016)}]{xgboost}
\bibinfo{author}{Chen, T.}, \bibinfo{author}{Guestrin, C.},
  \bibinfo{year}{2016}.
\newblock \bibinfo{title}{{XGBoost}: {A} {Scalable} {Tree} {Boosting}
  {System}}.
\newblock \bibinfo{journal}{Proceedings of the 22nd ACM SIGKDD International
  Conference on Knowledge Discovery and Data Mining - KDD '16} ,
  \bibinfo{pages}{785--794}.
\bibitem[{Covert et~al.(2020a)Covert, Lundberg and Lee}]{[S17]}
\bibinfo{author}{Covert, I.}, \bibinfo{author}{Lundberg, S.},
  \bibinfo{author}{Lee, S.I.}, \bibinfo{year}{2020}a.
\newblock \bibinfo{title}{Explaining by removing: A unified framework for model
  explanation}.
\newblock \href{http://arxiv.org/abs/2011.14878}{{\tt arXiv:2011.14878}}.
\bibitem[{Covert et~al.(2020b)Covert, Lundberg and
  Lee}]{covert2020understanding}
\bibinfo{author}{Covert, I.}, \bibinfo{author}{Lundberg, S.},
  \bibinfo{author}{Lee, S.I.}, \bibinfo{year}{2020}b.
\newblock \bibinfo{title}{Understanding global feature contributions with
  additive importance measures}.
\newblock \href{http://arxiv.org/abs/2004.00668}{{\tt arXiv:2004.00668}}.
\bibitem[{Efron(1987)}]{efron_better_1987}
\bibinfo{author}{Efron, B.}, \bibinfo{year}{1987}.
\newblock \bibinfo{title}{Better {Bootstrap} {Confidence} {Intervals}}.
\newblock \bibinfo{journal}{Journal of the American Statistical Association}
  \bibinfo{volume}{82}, \bibinfo{pages}{171--185}.
\bibitem[{Efron and Tibshirani(1994)}]{efron}
\bibinfo{author}{Efron, B.}, \bibinfo{author}{Tibshirani, R.J.},
  \bibinfo{year}{1994}.
\newblock \bibinfo{title}{{An Introduction to the Bootstrap}}.
\newblock \bibinfo{publisher}{Chapman \& Hall/CRC}.
\bibitem[{Fryer et~al.(2021a)Fryer, Strümke and Nguyen}]{shapley_good_bad}
\bibinfo{author}{Fryer, D.}, \bibinfo{author}{Strümke, I.},
  \bibinfo{author}{Nguyen, H.}, \bibinfo{year}{2021}a.
\newblock \bibinfo{title}{Shapley values for feature selection: The good, the
  bad, and the axioms}.
\newblock \href{http://arxiv.org/abs/2102.10936}{{\tt arXiv:2102.10936}}.
\bibitem[{Fryer et~al.(2021b)Fryer, Strumke and Nguyen}]{fryer2021model}
\bibinfo{author}{Fryer, D.V.}, \bibinfo{author}{Strumke, I.},
  \bibinfo{author}{Nguyen, H.}, \bibinfo{year}{2021}b.
\newblock \bibinfo{title}{Model independent feature attributions: Shapley
  values that uncover non-linear dependencies}.
\newblock \bibinfo{journal}{PeerJ Computer Science} \bibinfo{volume}{7},
  \bibinfo{pages}{e582}.
\bibitem[{Gagliano~Taliun et~al.(2020)Gagliano~Taliun, VandeHaar, Boughton,
  Welch, Taliun, Schmidt, Zhou, Nielsen, Willer, Lee, Fritsche, Boehnke and
  Abecasis}]{gagliano_taliun_exploring_2020}
\bibinfo{author}{Gagliano~Taliun, S.A.}, \bibinfo{author}{VandeHaar, P.},
  \bibinfo{author}{Boughton, A.P.}, \bibinfo{author}{Welch, R.P.},
  \bibinfo{author}{Taliun, D.}, \bibinfo{author}{Schmidt, E.M.},
  \bibinfo{author}{Zhou, W.}, \bibinfo{author}{Nielsen, J.B.},
  \bibinfo{author}{Willer, C.J.}, \bibinfo{author}{Lee, S.},
  \bibinfo{author}{Fritsche, L.G.}, \bibinfo{author}{Boehnke, M.},
  \bibinfo{author}{Abecasis, G.R.}, \bibinfo{year}{2020}.
\newblock \bibinfo{title}{Exploring and visualizing large-scale genetic
  associations by using {PheWeb}}.
\newblock \bibinfo{journal}{Nature Genetics} \bibinfo{volume}{52}.
\newblock \URLprefix \url{https://pheweb.org/UKB-TOPMed/}.
\bibitem[{Goeman and Solari(2014)}]{goeman_multiple_2014}
\bibinfo{author}{Goeman, J.J.}, \bibinfo{author}{Solari, A.},
  \bibinfo{year}{2014}.
\newblock \bibinfo{title}{Multiple hypothesis testing in genomics}.
\newblock \bibinfo{journal}{Statistics in Medicine} \bibinfo{volume}{33},
  \bibinfo{pages}{1946--1978}.
\bibitem[{Huettner and Sunder(2012)}]{Huettner:2012aa}
\bibinfo{author}{Huettner, F.}, \bibinfo{author}{Sunder, M.},
  \bibinfo{year}{2012}.
\newblock \bibinfo{title}{{Axiomatic arguments for decomposiing goodness of fit
  according to Shapley and Owen values}}.
\newblock \bibinfo{journal}{Electronic Journal of Statistics}
  \bibinfo{volume}{6}, \bibinfo{pages}{1239--1250}.
\bibitem[{Johnsen et~al.(2021)Johnsen, Riemer-Sørensen, DeWan, Cahill and
  Langaas}]{johnsen_new_2021}
\bibinfo{author}{Johnsen, P.V.}, \bibinfo{author}{Riemer-Sørensen, S.},
  \bibinfo{author}{DeWan, A.T.}, \bibinfo{author}{Cahill, M.E.},
  \bibinfo{author}{Langaas, M.}, \bibinfo{year}{2021}.
\newblock \bibinfo{title}{A new method for exploring gene–gene and
  gene–environment interactions in {GWAS} with tree ensemble methods and
  {SHAP} values}.
\newblock \bibinfo{journal}{BMC Bioinformatics} \bibinfo{volume}{22}.
\bibitem[{Karlsson et~al.(2019)Karlsson, Rask-Andersen, Pan, H{\"o}glund,
  Wadelius, Ek and Johansson}]{karlsson2019contribution}
\bibinfo{author}{Karlsson, T.}, \bibinfo{author}{Rask-Andersen, M.},
  \bibinfo{author}{Pan, G.}, \bibinfo{author}{H{\"o}glund, J.},
  \bibinfo{author}{Wadelius, C.}, \bibinfo{author}{Ek, W.E.},
  \bibinfo{author}{Johansson, {\AA}.}, \bibinfo{year}{2019}.
\newblock \bibinfo{title}{Contribution of genetics to visceral adiposity and
  its relation to cardiovascular and metabolic disease}.
\newblock \bibinfo{journal}{Nature medicine} \bibinfo{volume}{25},
  \bibinfo{pages}{1390--1395}.
\bibitem[{Karoui and Purdom(2018)}]{boothighdim}
\bibinfo{author}{Karoui, N.E.}, \bibinfo{author}{Purdom, E.},
  \bibinfo{year}{2018}.
\newblock \bibinfo{title}{Can {We} {Trust} the {Bootstrap} in
  {High}-dimensions? {The} {Case} of {Linear} {Models}}.
\newblock \bibinfo{journal}{Journal of Machine Learning Research}
  \bibinfo{volume}{19}, \bibinfo{pages}{66}.
\bibitem[{Keinan et~al.(2003)Keinan, Hilgetag, Meilijson and Ruppin}]{[F4]}
\bibinfo{author}{Keinan, A.}, \bibinfo{author}{Hilgetag, C.C.},
  \bibinfo{author}{Meilijson, I.}, \bibinfo{author}{Ruppin, E.},
  \bibinfo{year}{2003}.
\newblock \bibinfo{title}{Fair attribution of functional contribution in
  artificial and biological networks}.
\newblock \bibinfo{journal}{Neural Computation} \bibinfo{volume}{16},
  \bibinfo{pages}{1887--1915}.
\bibitem[{Kwon et~al.(2021)Kwon, Rivas and Zou}]{[S13]}
\bibinfo{author}{Kwon, Y.}, \bibinfo{author}{Rivas, M.A.},
  \bibinfo{author}{Zou, J.}, \bibinfo{year}{2021}.
\newblock \bibinfo{title}{Efficient computation and analysis of distributional
  {S}hapley values}.
\newblock \href{http://arxiv.org/abs/2007.01357}{{\tt arXiv:2007.01357}}.
\bibitem[{Locke et~al.(2015)Locke, Kahali, Berndt et~al.}]{locke_genetic_2015}
\bibinfo{author}{Locke, A.E.}, \bibinfo{author}{Kahali, B.},
  \bibinfo{author}{Berndt, S.I.}, et~al., \bibinfo{year}{2015}.
\newblock \bibinfo{title}{Genetic studies of body mass index yield new insights
  for obesity biology}.
\newblock \bibinfo{journal}{Nature} \bibinfo{volume}{518},
  \bibinfo{pages}{197--206}.
\bibitem[{Lundberg et~al.(2020)Lundberg, Erion, Chen, DeGrave, Prutkin, Nair,
  Katz, Himmelfarb, Bansal and Lee}]{lundberg_local_2020}
\bibinfo{author}{Lundberg, S.M.}, \bibinfo{author}{Erion, G.},
  \bibinfo{author}{Chen, H.}, \bibinfo{author}{DeGrave, A.},
  \bibinfo{author}{Prutkin, J.M.}, \bibinfo{author}{Nair, B.},
  \bibinfo{author}{Katz, R.}, \bibinfo{author}{Himmelfarb, J.},
  \bibinfo{author}{Bansal, N.}, \bibinfo{author}{Lee, S.I.},
  \bibinfo{year}{2020}.
\newblock \bibinfo{title}{From local explanations to global understanding with
  explainable {AI} for trees}.
\newblock \bibinfo{journal}{Nature Machine Intelligence} \bibinfo{volume}{2}.
\bibitem[{Lundberg et~al.(2019)Lundberg, Erion and Lee}]{[S11]}
\bibinfo{author}{Lundberg, S.M.}, \bibinfo{author}{Erion, G.G.},
  \bibinfo{author}{Lee, S.I.}, \bibinfo{year}{2019}.
\newblock \bibinfo{title}{Consistent individualized feature attribution for
  tree ensembles}.
\newblock \href{http://arxiv.org/abs/1802.03888}{{\tt arXiv:1802.03888}}.
\bibitem[{Lundberg and Lee(2017)}]{Lundberg2017}
\bibinfo{author}{Lundberg, S.M.}, \bibinfo{author}{Lee, S.I.},
  \bibinfo{year}{2017}.
\newblock \bibinfo{title}{{A Unified Approach to Interpreting Model
  Predictions}}, in: \bibinfo{editor}{Guyon, I.}, \bibinfo{editor}{Luxburg,
  U.V.}, \bibinfo{editor}{Bengio, S.}, \bibinfo{editor}{Wallach, H.},
  \bibinfo{editor}{Fergus, R.}, \bibinfo{editor}{Vishwanathan, S.},
  \bibinfo{editor}{Garnett, R.} (Eds.), \bibinfo{booktitle}{Advances in Neural
  Information Processing Systems 30}. \bibinfo{publisher}{Curran Associates,
  Inc.}, pp. \bibinfo{pages}{4765--4774}.
\bibitem[{Moehle et~al.(2021)Moehle, Boyd and Ang}]{[S16]}
\bibinfo{author}{Moehle, N.}, \bibinfo{author}{Boyd, S.}, \bibinfo{author}{Ang,
  A.}, \bibinfo{year}{2021}.
\newblock \bibinfo{title}{Portfolio performance attribution via {S}hapley
  value}.
\newblock \href{http://arxiv.org/abs/2102.05799}{{\tt arXiv:2102.05799}}.
\bibitem[{Redelmeier et~al.(2020)Redelmeier, Jullum and Aas}]{[S12]}
\bibinfo{author}{Redelmeier, A.}, \bibinfo{author}{Jullum, M.},
  \bibinfo{author}{Aas, K.}, \bibinfo{year}{2020}.
\newblock \bibinfo{title}{Explaining predictive models with mixed features
  using {S}hapley values and conditional inference trees}.
\newblock \href{http://arxiv.org/abs/2007.01027}{{\tt arXiv:2007.01027}}.
\bibitem[{Sellereite and Jullum(2019)}]{[S4]}
\bibinfo{author}{Sellereite, N.}, \bibinfo{author}{Jullum, M.},
  \bibinfo{year}{2019}.
\newblock \bibinfo{title}{shapr: An {R}-package for explaining machine learning
  models with dependence-aware {S}hapley values}.
\newblock \bibinfo{journal}{Journal of Open Source Software}
  \bibinfo{volume}{5}, \bibinfo{pages}{2027}.
\newblock \URLprefix \url{https://doi.org/10.21105/joss.02027},
  \DOIprefix\doi{10.21105/joss.02027}.
\bibitem[{{S}hapley(1953)}]{shapley_original}
\bibinfo{author}{{S}hapley, L.S.}, \bibinfo{year}{1953}.
\newblock \bibinfo{title}{A value for n-person games}, in:
  \bibinfo{booktitle}{Contributions to the Theory of Games (AM-28), Volume II}.
\bibitem[{Song et~al.(2016)Song, Nelson and Staum}]{[S15]}
\bibinfo{author}{Song, E.}, \bibinfo{author}{Nelson, B.},
  \bibinfo{author}{Staum, J.}, \bibinfo{year}{2016}.
\newblock \bibinfo{title}{{S}hapley effects for global sensitivity analysis:
  Theory and computation}.
\newblock \bibinfo{journal}{SIAM/ASA Journal on Uncertainty Quantification}
  \bibinfo{volume}{4}, \bibinfo{pages}{1060--1083}.
\newblock \DOIprefix\doi{10.1137/15M1048070}.
\bibitem[{Strumbelj and Kononenko(2010)}]{[S8]}
\bibinfo{author}{Strumbelj, E.}, \bibinfo{author}{Kononenko, I.},
  \bibinfo{year}{2010}.
\newblock \bibinfo{title}{An efficient explanation of individual
  classifications using game theory}.
\newblock \bibinfo{journal}{Journal of Machine Learning Research}
  \bibinfo{volume}{11}, \bibinfo{pages}{1--18}.
\newblock \DOIprefix\doi{10.1145/1756006.1756007}.
\bibitem[{Strumbelj and Kononenko(2013)}]{[S6]}
\bibinfo{author}{Strumbelj, E.}, \bibinfo{author}{Kononenko, I.},
  \bibinfo{year}{2013}.
\newblock \bibinfo{title}{Explaining prediction models and individual
  predictions with feature contributions}.
\newblock \bibinfo{journal}{Knowledge and Information Systems}
  \bibinfo{volume}{41}, \bibinfo{pages}{647--665}.
\newblock \DOIprefix\doi{10.1007/s10115-013-0679-x}.
\bibitem[{Sudlow et~al.(2015)Sudlow, Gallacher, Allen, Beral, Burton, Danesh,
  Downey, Elliott, Green, Landray et~al.}]{sudlow2015uk}
\bibinfo{author}{Sudlow, C.}, \bibinfo{author}{Gallacher, J.},
  \bibinfo{author}{Allen, N.}, \bibinfo{author}{Beral, V.},
  \bibinfo{author}{Burton, P.}, \bibinfo{author}{Danesh, J.},
  \bibinfo{author}{Downey, P.}, \bibinfo{author}{Elliott, P.},
  \bibinfo{author}{Green, J.}, \bibinfo{author}{Landray, M.}, et~al.,
  \bibinfo{year}{2015}.
\newblock \bibinfo{title}{Uk biobank: an open access resource for identifying
  the causes of a wide range of complex diseases of middle and old age}.
\newblock \bibinfo{journal}{PLoS medicine} \bibinfo{volume}{12},
  \bibinfo{pages}{e1001779}.
\bibitem[{Visscher et~al.(2017)Visscher, Wray, Zhang, Sklar, McCarthy, Brown
  and Yang}]{visscher_10_2017}
\bibinfo{author}{Visscher, P.M.}, \bibinfo{author}{Wray, N.R.},
  \bibinfo{author}{Zhang, Q.}, \bibinfo{author}{Sklar, P.},
  \bibinfo{author}{McCarthy, M.I.}, \bibinfo{author}{Brown, M.A.},
  \bibinfo{author}{Yang, J.}, \bibinfo{year}{2017}.
\newblock \bibinfo{title}{10 {Years} of {GWAS} {Discovery}: {Biology},
  {Function}, and {Translation}}.
\newblock \bibinfo{journal}{American Journal of Human Genetics}
  \bibinfo{volume}{101}, \bibinfo{pages}{5--22}.
\bibitem[{Young(1985)}]{Young:1985aa}
\bibinfo{author}{Young, H.P.}, \bibinfo{year}{1985}.
\newblock \bibinfo{title}{{Monotonic solutions of cooperative games}}.
\newblock \bibinfo{journal}{International Journal of Game Theory}
  \bibinfo{volume}{14}, \bibinfo{pages}{65--72}.
\bibitem[{Zhou et~al.(2018)Zhou, Nielsen, Fritsche, Dey, Gabrielsen, Wolford,
  LeFaive, VandeHaar, Gagliano, Gifford, Bastarache, Wei, Denny, Lin, Hveem,
  Kang, Abecasis, Willer and Lee}]{zhou_efficiently_2018}
\bibinfo{author}{Zhou, W.}, \bibinfo{author}{Nielsen, J.B.},
  \bibinfo{author}{Fritsche, L.G.}, \bibinfo{author}{Dey, R.},
  \bibinfo{author}{Gabrielsen, M.E.}, \bibinfo{author}{Wolford, B.N.},
  \bibinfo{author}{LeFaive, J.}, \bibinfo{author}{VandeHaar, P.},
  \bibinfo{author}{Gagliano, S.A.}, \bibinfo{author}{Gifford, A.},
  \bibinfo{author}{Bastarache, L.A.}, \bibinfo{author}{Wei, W.Q.},
  \bibinfo{author}{Denny, J.C.}, \bibinfo{author}{Lin, M.},
  \bibinfo{author}{Hveem, K.}, \bibinfo{author}{Kang, H.M.},
  \bibinfo{author}{Abecasis, G.R.}, \bibinfo{author}{Willer, C.J.},
  \bibinfo{author}{Lee, S.}, \bibinfo{year}{2018}.
\newblock \bibinfo{title}{Efficiently controlling for case-control imbalance
  and sample relatedness in large-scale genetic association studies}.
\newblock \bibinfo{journal}{Nature Genetics} \bibinfo{volume}{50}.

\end{thebibliography}

\end{document}